\documentclass[journal]{IEEEtran}

\usepackage{amsmath}
\usepackage{amsfonts}
\usepackage{amssymb}
\usepackage{microtype}
\usepackage{bm}
\usepackage{algorithm}
\usepackage{algorithmic}  
\newcommand{\algorithmicinput}{\textbf{Input:}} 
\newcommand{\algorithmicoutput}{\textbf{Output:}}
\usepackage{amsmath} 
\usepackage{color}
\usepackage{float}
\usepackage{booktabs}
\usepackage{multirow}
\usepackage{graphicx}
\usepackage{placeins}
\usepackage{cite}
\usepackage{footnote}
\usepackage{booktabs}
\usepackage{stfloats}
\usepackage{verbatim}
\usepackage{balance}
\usepackage{url}
\usepackage{epstopdf}
\usepackage[caption=false,font=normalsize,labelfont=sf,textfont=sf]{subfig}
\usepackage[hypertexnames=false, colorlinks=true, linkcolor=blue, citecolor=blue, urlcolor=cyan]{hyperref}
\def\BibTeX{{\rm B\kern-.05em{\sc i\kern-.025em b}\kern-.08em
		T\kern-.1667em\lower.7ex\hbox{E}\kern-.125emX}}

\ifCLASSINFOpdf
 
\else
 
\fi

\begin{document}

\title{Multi-view Clustering via Unified Multi-kernel Learning and Matrix Factorization}

\author{Chenxing Jia,
        Mingjie Cai,
        Hamido Fujita 
\thanks{This work was supported in part by the National Natural Science Foundation of China under Project 12471431, Project 12231007, in part by the  Hunan Provincial Natural Science Foundation of China under Project 2023JJ30113 and in part by the Guangdong Basic and Applied Basic Research Foundation under Project 2023A1515012342. \textit{(Corresponding author: Mingjie Cai.)}} 
\thanks{Chenxing Jia is with School of Mathematics, Hunan University, Hunan, Changsha 410082, China (e-mail: Jcx\_hnu@163.com).}
\thanks{Mingjie Cai is with School of Mathematics, Hunan University, Hunan, Changsha 410082, China; Shenzhen Research Institute, Hunan University, Shenzhen, Guangdong 518000, China; and Hunan Key Laboratory of Intelligent Decision-Making Technology for Emergency Management, Hunan University, Hunan, Changsha 410082, China (e-mail: cmjlong@163.com).}
\thanks{Hamido Fujita is with Malaysia-Japan International Institute of Technology (MJIIT), Universiti Teknologi, Malaysia, 54100 Kuala Lumpur, Malaysia; College of Science, Princess Nourah bint Abdulrahman University (PNU), 13412 Riyadh, Saudi Arabia; and Regional Research Center, Iwate Prefectural University, Iwate 0200693 Japan (e-mail: hfujita-799@acm.org).}
}

\maketitle

\begin{abstract}
Multi-view clustering has become increasingly important due to the multi-source character of real-world data. Among existing multi-view clustering methods, multi-kernel clustering and matrix factorization-based multi-view clustering have gained widespread attention as mainstream approaches. However, multi-kernel clustering tends to learn an optimal kernel and then perform eigenvalue decomposition on it, which leads to high computational complexity. Matrix factorization-based multi-view clustering methods impose orthogonal constraints on individual views. This overly emphasizes the accuracy of clustering structures within single views and restricts the learning of individual views. Based on this analysis, we propose a multi-view clustering method that integrates multi-kernel learning with matrix factorization. This approach combines the advantages of both multi-kernel learning and matrix factorization. It removes the orthogonal constraints on individual views and imposes orthogonal constraints on the consensus matrix, resulting in an accurate final clustering structure. Ultimately, the method is unified into a simple form of multi-kernel clustering, but avoids learning an optimal kernel, thus reducing the time complexity. Furthermore, we propose an efficient three-step optimization algorithm to achieve a locally optimal solution. Experiments on widely-used real-world datasets demonstrate the effectiveness of our proposed method.
\end{abstract}


\IEEEpeerreviewmaketitle

\section{Introduction}

\IEEEPARstart{M}{ulti-view} clustering (MVC) is an unsupervised learning method designed to integrate information from multiple distinct feature sets or data sources\cite{wangStudyGraphbasedSystem2019}, \cite{xijiongxieMultiviewClusteringEnsembles2013}, \cite{yaoMultipleKernelMeans2021}, \cite{Wangmultiviewfuzzy2022}. In real-world applications, data is often multi-dimensional and multi-source\cite{LinConsistentgraph2023}, \cite{gaoMultiviewSubspaceClustering2015}, \cite{LINMultiviewclustering2023}, \cite{wangMultiviewFuzzyClustering2017}. For example, in image recognition tasks, the same image can be described by different features such as color, texture, and shape. In bioinformatics, the same biological entity can be described in various ways, such as through gene expression, protein sequences, and cellular functions. These multiple descriptions of the same object represent observations from different views, which may contain complementary information, allowing for a more comprehensive characterization of the object, thus leading to more accurate clustering\cite{huangFastMultiViewClustering2023},\cite{jiangFastMultipleGraphs2022}, \cite{liFlexibleMultiViewRepresentation2019}, \cite{liuClusterWeightedKernelKMeans2020}. Multi-view clustering can extract richer and more reliable information than single-view methods, which is why it has gained the attention of many researchers.

Existing multi-view clustering methods can be roughly divided into four categories: deep multi-view clustering (DMVC)\cite{duDeepMultipleAutoEncoderBased2021}, \cite{wangTripleGranularityContrastiveLearning2023}, \cite{FangDMRLNetDifferentiable2023}, multi-view subspace clustering (MVSC)\cite{yinMultiviewSubspaceClustering2019}, \cite{FuUnifiedLowRank2023}, multiple kernel clustering (MKC) \cite{liangConsistencyLargeScaleExtension2024}, \cite{liLocalSampleWeightedMultiple2022}, \cite{wangMultipleKernelClustering2024}, \cite{liuLocalizedSimpleMultiple2021}, \cite{wangMultiviewClusteringLate2019}, and matrix factorization-based methods (MFMVC)\cite{liuMultiViewNonnegativeMatrix2020}, \cite{wanOneStepMultiViewClustering2024a}, \cite{zhangConsensusOneStepMultiView2022}, \cite{zhangMultiviewClusteringMultimanifold2014}. The latter three can also be referred to as heuristic learning. DMVC utilize multilayer mappings of neural networks to achieve the desired clustering effects, but the networks constructed are often complex and difficult to interpret.

The key to MVSC lies in constructing affinity graphs\cite{liuEfficientOnePassMultiView2022}, where most affinity graphs can be built based on a self-expression mechanism, believing that each sample can be expressed linearly by other samples. To ensure that the affinity matrix reflects the global structure of the data for subsequent clustering, some researchers incorporate various regularization constraints, such as sparsity\cite{liuMultiviewSubspaceClustering2022} and low-rank\cite{lanMultiviewSubspaceClustering2024} constraints, to solve for the affinity matrix.

In MKC, a series of pre-defined kernel matrices are typically used to capture inter-view information. For example, baseline approach multiple kernel k-means (MKKM)\cite{hsin-chienhuangMultipleKernelFuzzy2012} employs a linear combination to obtain the optimal kernel, and subsequent variants are generally based on the assumption that the linear combination of base kernels is the optimal kernel. Similar to affinity matrix learning, many MKC methods also introduce constraints on the kernels to achieve optimal results. For instance, Li et al.\cite{liMultipleKernelClustering2016} optimizes the similarity between samples and their k-nearest neighbors through local kernel alignment, while \cite{liuMultipleKernelKMeans2016} introduces matrix-induced regularization to reduce redundancy among selected kernels. kernel coefficients also can be optimized by maximizing the alignment between the combined kernel and the ideal target kernel\cite{luMultipleKernelClustering2014}. 
MKC usually alternates between optimizing the clustering indicator matrix and view coefficients to find local minima. Liu et al.\cite{liuSimpleMKKMSimpleMultiple2023} have transformed the objective function resolution method to achieve global optimal solutions. Despite the significant progress made by the aforementioned kernel methods, their ultimate form is still based on the decomposition of features into a specific optimal kernel, which is generally a linear combination of pre-processed kernels or neighborhood kernels of linear kernels. The handling of the optimal kernel typically involves high computational complexity.

\begin{figure*}[!ht]
	\centering
	\includegraphics[width=\textwidth]{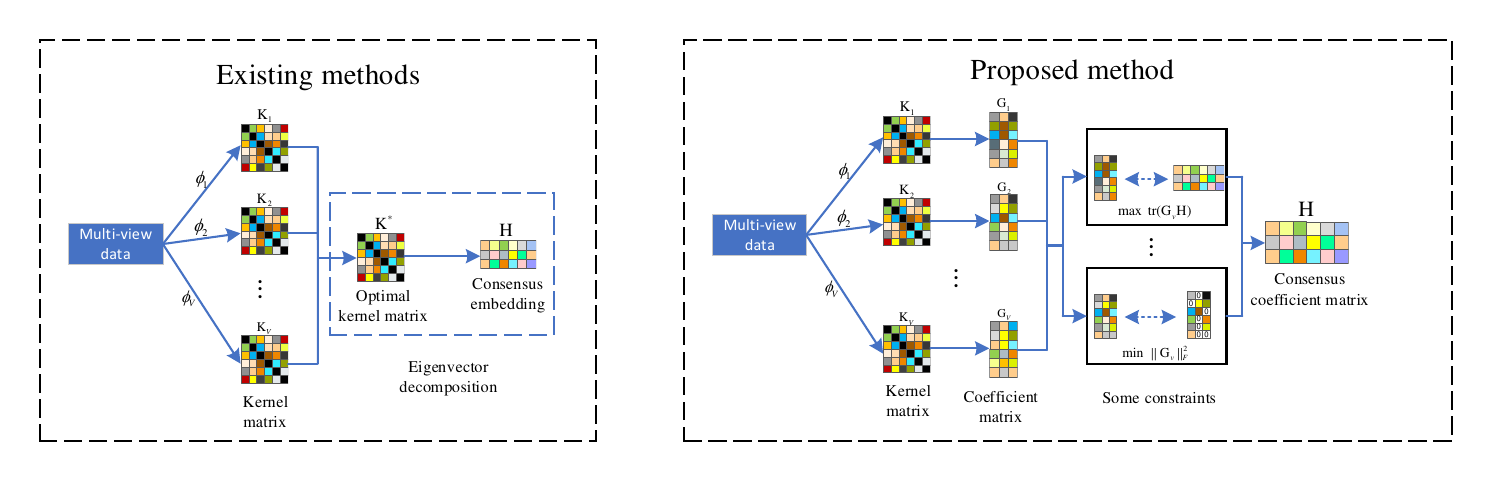} 
	\caption{Comparison between existing kernel-based methods and the proposed method. Existing methods typically learn an optimal kernel from the kernel matrices of multiple views, followed by direct eigenvector decomposition to obtain $\mathbf{H}$. However, the introduction of the optimal kernel results in high computational complexity. In contrast, the proposed method directly performs matrix factorization on the kernel matrix of each view under constraints such as sparsity, thereby learning a consensus coefficient matrix $\mathbf{H}$.}
	\label{fig1}
\end{figure*}

In contrast, MFMVC tend to have lower time complexity. Classic matrix factorization for clustering is Non-negative Matrix Factorization (NMF), which decomposes the data matrix into two non-negative parts, referred to as the basis matrix and coefficient matrix. However, the non-negativity constraint can somewhat limit the scope of learning, leading some researchers\cite{liuOnepassMultiviewClustering2021} to remove this constraint. Additionally, in multi-view settings, most researchers focus on learning a consensus coefficient matrix, such as directly obtaining the average of each view\cite{wangDiverseNonNegativeMatrix2018} or learning the consensus matrix directly\cite{liuMultiViewClusteringJoint2013}. However, these methods often impose orthogonality constraints on the basis matrix of single views during the learning process, which may not be reasonable since the information from single views is not complete or reliable in a multi-view context. Wan et al.\cite{wanAutoweightedMultiviewClustering2023} has recognized this issue and considered different dimensions, but different dimensions remain limited. Thus, is such an orthogonality constraint still too stringent?

Based on the above analysis, this article proposes a method for multi-view clustering via unified multi-kernel learning and matrix factorization (MVC-UMKLMF) with the aim of combining the advantages of MKC and MFMVC while addressing the aforementioned shortcomings. MVC-UMKLMF first considers matrix factorization of the data matrices after mapping through multiple view kernel functions, removing the non-negativity constraint. At the same time, it eliminates the orthogonality constraint on the basis matrix of individual views and directly imposes the orthogonality constraint on the consensus coefficient matrix, which can be understood as the clustering indicator matrix. The final form of MVC-UMKLMF integrates into a kernel matrix-based objective function, achieving a formal unification of MKC and MFMVC. Additionally, we propose a three-step alternating optimization algorithm to solve the proposed objective function, ensuring that each optimization step leads to a monotonic decrease in the objective function, reaching a local optimal solution, while reducing the time complexity compared to MKC.

The main contributions of this paper are summarized as follows:

1) The removal of the non-negativity constraint in matrix decomposition and the orthogonality constraint on the basis matrix, while directly imposing the orthogonality constraint on the consensus matrix, allows for more flexibility in learning the clustering structure of individual views.

2) Mapping the data matrices of each view through kernel functions enables the handling of nonlinear data and avoids the computationally intensive steps of learning the optimal kernel and eigenvector decomposition.

3) The unification of matrix factorization and MKC is achieved, along with the presentation of a simple three-step optimization algorithm, MVC-UMKLMF.

The remaining structure of the paper is arranged as follows. Section~\ref{sec2} introduces the related work. Section~\ref{sec3} presents the motivation for our work, the objective function, and the solution algorithm. Section~\ref{sec4} provides a theoretical analysis of the proposed work, including convergence analysis and time complexity analysis. Section~\ref{sec5} demonstrates the effectiveness of the proposed method through comparative experiments. Section~\ref{sec6} concludes the paper.

\section{Related Work} \label{sec2}
\subsection{NMF for clustering}
Given a dataset $\mathbf{X}=[\mathbf{x_1},\mathbf{x_2},...,\mathbf{x_n}]\in {\mathbb{R}^{d\times n}}$, NMF aims to decompose $\mathbf{X}$ into the product of two non-negative matrices:
\begin{equation}
\label{eq.1}
    \underset {\mathbf{H}\geqslant 0,\mathbf{Z}\geqslant 0} {\min}l(\mathbf{X},\mathbf{Z}\mathbf{H})
\end{equation}
in which $\mathbf{Z}\in{\mathbb{R}^{d\times k}}$ is the basis matrix, $\mathbf{H}\in{\mathbb{R}^{k\times n}}$ is the coefficient matrix, $k$ is the desired embedding dimension, and $l\left ( \cdot ,\cdot  \right )$ represents the reconstruction loss between $\mathbf{X}$ and $\mathbf{Z}\mathbf{H}$. Considering NMF for clustering,  to facilitate the interpretation of the clustering, an orthogonal constraint is generally imposed on $\mathbf{Z}$, assuming that the basis vectors in the basis matrix should be mutually orthogonal\cite{wanOneStepMultiViewClustering2024a}. The form is as follows:
    \begin{equation}
    \label{eq.2}
    	\begin{aligned}
    		\underset {\mathbf{H}, \mathbf{Z}}\min &\quad l\left(\mathbf{X},\mathbf{Z}\mathbf{H}\right)+\Psi \left (\mathbf{H}, \mathbf{Z} \right )\\
    		s.t.&\quad \mathbf{H}\geqslant 0, \mathbf{Z}\geqslant 0, \mathbf{Z}^T\mathbf{Z}=\mathbf{I_k} \\
    	\end{aligned}
    \end{equation}
	in which $\Psi \left ( \cdot ,\cdot  \right )$ is a regularization term. Ding et al.\cite{dingOrthogonalNonnegativeMatrix2006} demonstrated that imposing an orthogonal constraint on $\mathbf{H}$ is equivalent to k-means.
\subsection{Multi-View Clustering Based on Matrix Factorization}
Eq. (\ref{eq.2}) generalizes to multi-view by naturally replacing $\mathbf{X}$, $\mathbf{H}$, $\mathbf{Z}$ with $\mathbf{Z}_v$, $\mathbf{H}_v$, $ \mathbf{Z}_v$ respectively. Given multi-view data $\mathbf{X}=\left\{ \mathbf{X}_{1},\mathbf{X}_{2},...,\mathbf{X}_{V}\right\}$, replace $l\left ( \cdot ,\cdot  \right )$ with the common used Frobenius norm,
similar to single-view, one of the basic models of MVCMF is to directly apply matrix factorization to each view, and then fuse the multiple coefficient matrices. The optimization function is as follows :
        \begin{equation}
        \label{eq.3}
    		\begin{aligned}
    			\underset {\left \{ \mathbf{H}_v \right \} ,\left \{ \mathbf{Z}_v \right \}}\min& \quad \sum_{v=1}^{V}\left[\left \| \mathbf{X}_v-\mathbf{Z}_v\mathbf{H}_v \right \|_{F}^{2}+f\left(\mathbf{Z}_v,\mathbf{H}_v\right)\right]\\
    			s.t.&\quad \mathbf{Z}_v\geqslant 0, \mathbf{H}_v\geqslant 0\\
    		\end{aligned}
    	\end{equation}
where $f\left ( \cdot ,\cdot  \right )$ represents additional processing, such as regularization terms or considering redundant information. The average of the coefficient matrices from each view, $\mathbf{H}^{*}=\sum_{v=1}^{V} \mathbf{H}_v$, is regarded as the final learned consensus coefficient matrix\cite{wangDiverseNonNegativeMatrix2018}.

Another common model is to directly learn the consensus coefficient matrix from the multi-view matrix factorization, eliminating the need for a fusion step, as follows\cite{DataFusionMetabolomics}:
            \begin{equation}
            \label{eq.4}
    		\begin{aligned}
    			\underset{ \mathbf{H} ,\left \{ \mathbf{Z}_v \right \}}\min &\quad \sum_{v=1}^{V} \left \| \mathbf{X}_v-\mathbf{Z}_v\mathbf{H} \right \|_{F}^{2}\\
    			s.t.&\quad \mathbf{H}\geqslant 0, \mathbf{Z}_v\geqslant 0, \mathbf{Z}_v^T\mathbf{Z}_v = \mathbf{I}_k\\    			
    		\end{aligned}
    	\end{equation}\\
     or, as shown below, directly learn the coefficient matrices and consensus matrix of each view in the objective function at the same time\cite{liuMultiViewClusteringJoint2013}:

  \begin{equation}
  \label{eq.5}
    		\begin{aligned}
    			\underset{ \left\{\mathbf{H}_v\right\},\left \{ \mathbf{Z}_v \right \},\mathbf{H}^*}\min &\quad\sum_{v=1}^{V}\left[ \left \| \mathbf{X}_v-\mathbf{Z}_v\mathbf{H}_v \right \|_{F}^{2}+\left\|\mathbf{H}_v-\mathbf{H}^*\right\|_{F}^{2}\right]\\  
    			s.t.&\quad \mathbf{H}\geqslant 0,\mathbf{Z}_v\geqslant 0,\mathbf{Z}_v^T\mathbf{Z}_v = \mathbf{I}_k.\\    			
    		\end{aligned}
    	\end{equation}
        
Both of the above basic models feed the learned consensus coefficient matrix $\mathbf{H}$ into subsequent clustering steps to obtain results, typically by directly applying k-means to $\mathbf{H}$. Liu et al.\cite{liuOnepassMultiviewClustering2021} removes the non-negativity constraint to strive for a more relaxed learning scope.
\subsection{Multiple Kernel K-means}
We first review the objective function of kernel k-means (KKM), which is written in the following form:
        \begin{equation}
        \label{eq.6}
        	\underset {\mathbf{H}}\min\quad tr\left ( \mathbf{K}-\mathbf{H}\mathbf{K}\mathbf{H}^{T}   \right ) \quad
        	s.t.\quad \mathbf{H}\mathbf{H}^{T}=\mathbf{I}_{k}	
        \end{equation}
where $\mathbf{K}=\psi \left( \mathbf{X}
 \right)^{T} \psi \left( \mathbf{X} \right)$ is the precomputed kernel matrix, $\mathbf{H}$ is the clustering indicator matrix. In the multi-view setting, KKM is extended to MKKM, where the optimal kernel is considered to be a linear combination of base kernels:
    \begin{equation}
    \label{eq.7}
    	\underset{\mathbf{H}}\min\quad tr\left(\mathbf{K}_{\mathbf{\gamma}  }-\mathbf{H}\mathbf{K}_{\mathbf{\gamma }}\mathbf{H}^{T}\right)\quad s.t.\quad \mathbf{H}\mathbf{H}^{T}=\mathbf{I}_{k}
    \end{equation}
in which the optimal kernel $\mathbf{K}_{\mathbf{\gamma} }=\sum_{v=1}^{V}\gamma_{v}^{2}\mathbf{K}_{v}$, where $\mathbf{K}_{v}$ is the kernel matrix for the $v$-th view. This optimization problem typically uses an alternating optimization method, alternatingly updating $\mathbf{H}$ and $\mathbf{\gamma}$ until convergence criteria are met. Many variants have been proposed based on MKKM\cite{liuSimpleMKKMSimpleMultiple2023, liuLocalizedSimpleMultiple2021}, but most assume that the optimal kernel is a linear combination of multiple kernels. Other algorithms\cite{liuOptimalNeighborhoodKernel2017} learn the optimal kernel $\mathbf{K}^{*}$ from the neighborhood of linear kernels, and ultimately feed the optimal kernel into KKM.

\section{Proposed method} \label{sec3}
\subsection{Motivation}
Kernel functions map multi-view data into a high-dimensional space, with the expectation that the data will exhibit linear characteristics in this space, thus enabling the handling of non-linear data. From the perspective of kernel-based multi-view clustering, existing methods are primarily based on the assumption that the optimal kernel is a linear combination of kernels from each view, or they learn an optimal kernel from the neighborhood of linear kernels, and then implement KKM based on this optimal kernel. From the perspective of matrix factorization-based multi-view clustering, existing work generally imposes orthogonality on  $\mathbf{Z}^{(v)}$  in Eq. (\ref{eq.4}) to have good clustering interpret-ability. Reviewing the related work, we raise three questions about this:

1) Is the introduction of the optimal kernel necessary in multi-kernel learning, and is it possible to directly learn the clustering consensus matrix from multiple kernels?

2) In Eq. (\ref{eq.5}), $\mathbf{Z}^{(v)}$ has a fixed dimension, and the clustering structure of multi-view data may not be clear under a single view. Is the orthogonality imposed on it too stringent?

3) The computational complexity of kernel-based methods and graph-based methods for multi-view clustering is generally O($n^3$). Can we seek techniques with lower computational complexity?

\subsection{Proposed Formula}
The above questions have led us to connect matrix factorization techniques with kernel-based multi-view clustering. In existing work, Tolic et al.\cite{tolicNonlinearOrthogonalNonNegative2018} first used kernel orthogonal NMF for non-negative spectral clustering. Inspired by this, for multi-view data $\mathbf{X}=\left\{ \mathbf{X}_{1},\mathbf{X}_{2},...,\mathbf{X}_{V}\right\}$, we use different kernel functions to map each view to a high-dimensional space, and directly learn a consensus embedding 
$\mathbf{H}$ in the multi-view high-dimensional space, represented in the following form:
    \begin{equation}
    \label{eq.8}
    	\begin{aligned}
    	\underset{\{W_{v}\},\mathbf{H}}\min&\quad \sum_{v=1}^{V} \left \| \phi_{v}\left ( \mathbf{X}_{v} \right )-\mathbf{W}_{v}\mathbf{H}  \right \|_{F}^{2}\\ 
    	s.t.&\quad \mathbf{H}\mathbf{H}^{T}=\mathbf{I}_{k}
        \end{aligned}
    \end{equation}
in which $\mathbf{W}_{v}\in{\mathbb{R}^{d_{v}^{'}\times k}}$ is the basis matrix for view $v$, and due to the differences among various views, we learn different basis matrices. These different basis matrices represent the distinct information of each view to some extent. $\mathbf{H} \in {\mathbb{R}^{k \times n}}$ is the consensus ecoefficient matrix, which is finally input into k-means to obtain clustering results, so it can also be referred to as consensus embedding. It is worth noting that although Eq. (\ref{eq.8}) exhibits the form of NMF, we remove the non-negativity constraint in order to learn the optimal embedding within a broader range. Furthermore, following\cite{dingConvexSemiNonnegativeMatrix2010}, we consider the basis matrix to be a linear combination of high-dimensional data points, thus $W_{v} = \phi_{v} \left(\mathbf{X}_{v}\right)\mathbf{G}_{v}$, and Eq. (\ref{eq.8}) is equivalent to:
    \begin{equation}
    \label{eq.9}
    	\begin{aligned}
    		\underset{\{\mathbf{G}_{v}\},\mathbf{H}}\min&\quad \sum_{v=1}^{V} \left \| \phi_{v}\left ( \mathbf{X}_{v} \right )- \phi_{v} \left(\mathbf{X}_{v}\right)\mathbf{G}_{v}\mathbf{H}  \right \|_{F}^{2}\\ 
    		s.t.&\quad \mathbf{H}\mathbf{H}^{T}=\mathbf{I}_{k}.
    	\end{aligned}
    \end{equation}

In practical operations, $\phi \left(\mathbf{X}\right)$  is high-dimensional, making it difficult to optimize Eq. (\ref{eq.9}). According to $\phi^T \left(\mathbf{x}_i\right) \phi \left(\mathbf{x}_j\right) = \mathbf{K}(x_i,x_j)$, we hope it can be transformed into an optimization formula based on the kernel matrix. Therefore, we expand it as follows:
    \begin{equation}
    \label{eq.10}
    	\begin{aligned}
    		&\sum_{v=1}^{V} \left \| \phi_{v}\left ( \mathbf{X}_{v} \right )- \phi_{v} \left(\mathbf{X}_{v}\right)\mathbf{G}_{v}\mathbf{H}  \right \|_{F}^{2}\\ 
    		=&\sum_{v=1}^{V} tr\left [ \phi_{v}^{T}\left ( \mathbf{X}_{v} \right )\phi_{v}\left ( \mathbf{X}_{v} \right )-2\phi_{v}^{T}\left ( \mathbf{X}_{v} \right )\phi_{v}\left ( \mathbf{X}_{v} \right )\mathbf{G}_{v}\mathbf{H}\right] \\
      &+\sum_{v=1}^{V} tr\left [\mathbf{H}^{T}\mathbf{G}_{v}^{T}\phi_{v}^{T}\left ( \mathbf{X}_{v} \right )\phi_{v}\left ( \mathbf{X}_{v} \right )\mathbf{G}_{v}\mathbf{H} \right ]\\
    		=&\sum_{v=1}^{V} tr\left [ \mathbf{K}_v-2\mathbf{K}_v\mathbf{G}_{v}\mathbf{H}+\mathbf{H}^{T}\mathbf{G}_{v}^{T}\phi_{v}^{T}\left ( \mathbf{X}_{v} \right )\phi_{v}\left ( \mathbf{X}_{v} \right )\mathbf{G}_{v}\mathbf{H} \right ].
    	\end{aligned}
    \end{equation}

The first term is the precomputed kernel matrix, which is a constant. Furthermore, due to the orthogonality of $\mathbf{H}$, the formula is equivalent to:
    \begin{equation}
    \label{eq.11}
    	\begin{aligned}
    	\underset{\mathbf{H},\{\mathbf{G}_{v}\}}\min &\quad\sum_{v=1}^{V} tr\left[-2\mathbf{H}\mathbf{K}_{v}\mathbf{G}_{v}+\mathbf{G}_{v}^{T}\mathbf{K}_{v}\mathbf{G}_{v}\right]\\
    	s.t.&\quad \mathbf{H}\mathbf{H}^{T}=I_{k}.    	
        \end{aligned}
    \end{equation}

Thus, Eq. (\ref{eq.11}) has been transformed in form, becoming a problem of maximizing the trace. Here, $\mathbf{K}_v$  is the precomputed kernel matrix for the $v$-th view, and $\mathbf{G}_v \in{\mathbb{R}^{n \times k}}$ represents the relationship between high-dimensional samples and the basis matrix in the $v$-th view. Since the basis matrix has a fixed dimension of $k$, the clustering result in a single view is not clearly defined as $k$ clusters, hence we do not require $\mathbf{G}_{v}$ to be orthogonal, leaving room for individual views. The consensus coefficient matrix $\mathbf{H}$ is directly obtained from the multi-view kernel matrices, avoiding the introduction of an optimal kernel.

Additionally, $\mathbf{H}$ as the consensus coefficient matrix and $\mathbf{G}_v$ as the coefficient matrix for a single view naturally have a close relationship. It is evident from Eq. (\ref{eq.9}) that the equation $\mathbf{G}_{v}\mathbf{H}=\mathbf{I}_k$ is expected to be satisfied. Multiplying both sides of this equation by $\mathbf{H}^{T}$ on the right implies that $\mathbf{G}_{v}=\mathbf{H}^T$, which provides a more direct guidance for the learning of $\mathbf{H}$. Therefore, we make a change to the above equation, and the new formula is as follows:
    \begin{equation}
    \label{eq.12}
    	\begin{aligned}
    		\underset{\mathbf{H},\{\mathbf{G}_{v}\}}\max &\quad \sum_{v=1}^{V} \left[tr\left(\mathbf{H}\mathbf{K}_{v}\mathbf{G}_{v}\right)+tr\left(\mathbf{G}_v\mathbf{H}\right)\right] \\
    		s.t.&\quad \mathbf{H}\mathbf{H}^{T}=\mathbf{I}_{k}
    	\end{aligned}
    \end{equation}
in which the first term results from replacing $\mathbf{G}_{v}^{T}$ in the second term of Eq. (\ref{eq.11}) with $\mathbf{H}$ and merging it with the first term, and the second term serves as a complement to the replacement, forcing $\mathbf{G}_{v}$ to be as close as possible to $\mathbf{H}^{T}$. Since the importance of each view varies, we automatically weight them and consider the second term separately, adding a parameter to control it. Thus, the final objective function after the transformation is as follows:
    \begin{equation}
    \label{eq.13}
   	\begin{aligned}
   		\underset{\mathbf{H},\{\mathbf{G}_{v}\}}\max&\quad \sum_{v=1}^{V}\omega_{v}^{2}\left [ tr\left(\mathbf{H}\mathbf{K}_{v}\mathbf{G}_{v}\right)+\alpha tr\left(\mathbf{G}_v\mathbf{H}\right)\right ]  \\
   		s.t.&\quad \mathbf{H}\mathbf{H}^{T}=\mathbf{I}_{k},\bm{\omega}\mathbf{1}=1.
   	\end{aligned}
   \end{equation}

\subsection{The Differences From Existing Methods}
Fig \ref{fig1} illustrates the difference between the final form of our proposed method and kernel-based methods. Focusing solely on the final objective form while ignoring the details of the derivation, work similar to this paper is derived from MKKM, with existing efforts focusing on learning an optimal kernel and then applying it to KKM to obtain the final clustering results. However, this work does not aim to learn an optimal kernel but instead directly learns a consensus embedding through multi-kernel learning. Starting from matrix factorization in high-dimensional space, we provide a new interpretation, remove the non-negativity constraint, and generalize it to multi-view data while being able to handle non-linear data.

\subsection{Optimization}
For the sake of optimization, and because the clustering structure is as clear as possible, $\mathbf{G}_v$ should be sparse. On the basis of Eq. (\ref{eq.13}), we add a regularization term for $\mathbf{G}_v$, thus the optimization objective formula is equivalent to:
   \begin{equation}
   \label{eq.14}
   	\begin{aligned}
   		\underset{\mathbf{H},\mathbf{G}_{v}}\max&\quad \sum_{v=1}^{V}\omega_{v}^{2}\left [ \left \| \mathbf{K}_{v}-\mathbf{G}_{v} \mathbf{H}\right \|_{F}^{2}+\alpha \left \| \mathbf{G}_{v}-\mathbf{H}^T \right \|_{F}^{2}\right ]  \\
   		s.t.&\quad \mathbf{H}\mathbf{H}^{T}=I_{k}, \bm{\omega}\mathbf{1}=1
   	\end{aligned}
   \end{equation}
which involves three variables $\mathbf{H}$, $\{\mathbf{G}_v\}$, $ \bm{\omega}$, we adopt the alternating optimization method, fixing two variables each time to optimize the other one. Therefore, this optimization problem is divided into three sub-problems.

\textit{1) Subproblem of $\{\mathbf{G}_v\}$}

Expanding Eq. (\ref{eq.14}) and removing terms that are independent of $\mathbf{G}_v$, Eq. (\ref{eq.14}) can be rewritten in the following form:
\begin{equation}
\label{eq.15}
   \begin{aligned}
   J = \left(\alpha+1\right)tr\left(\mathbf{G}_{v}^T\mathbf{G}_v\right)&-2\alpha tr\left(\mathbf{G}_{v}\mathbf{H}\right)\\&-2tr\left(\mathbf{K}_{v}^{T}\mathbf{H}^T\mathbf{G}_{v}^T\right).
   \end{aligned}
\end{equation}
Take the partial derivative of $J$ with respect to $\mathbf{G}_v$, $\frac{\partial J}{\partial \mathbf{G}_v} = 2\left(\alpha+1\right)\mathbf{G}_v-2\mathbf{K}_v^T\mathbf{H}^T-2\alpha \mathbf{H}^T$, set it to zero, then 

\begin{equation}
\label{eq.16}
\mathbf{G}_v=\frac{1}{ \left(\alpha+1\right)} \mathbf{K}_v^T\mathbf{H}^T+\frac{\alpha}{ \left(\alpha+1\right)} \mathbf{H}^T.
\end{equation}

\textit{2) Subproblem of $\mathbf{H}$}

With $\mathbf{G}_v$ and $\bm{\omega}$ fixed, equation (14) expands and is rewritten as:
\begin{equation}
\label{eq.17}
	\begin{aligned}
	\underset{\mathbf{H}}\max&\quad \sum_{v=1}^{V}w_v^2 \left[tr\left(\mathbf{K}_v^{T}\mathbf{H}^T\mathbf{G}_v^T\right)+tr\left(\mathbf{G}_v\mathbf{H}\right)\right]\\
	s.t.&\quad \mathbf{H}\mathbf{H}^{T}=\mathbf{I}_{k}.
    \end{aligned}
\end{equation}

Eq. (\ref{eq.17}) is equivalent to:
\begin{equation}
\label{eq.18}
	\begin{aligned}
		\underset{\mathbf{H}}\max&\quad tr\left( \mathbf{H}^T\mathbf{A}  \right) \\
		s.t.&\quad \mathbf{H}\mathbf{H}^{T}=I_{k}
	\end{aligned}
\end{equation}
where $\mathbf{A} = \sum_{v=1}^{V}w_v^2 \left ( \mathbf{G}_v^T\mathbf{K}_v^{T}+\mathbf{G}_v^T \right )$, the problem can be easily solved using SVD (Singular Value Decomposition). Performing SVD on $\mathbf{A}$ to obtain $\mathbf{A}=U \Sigma V^T$, then
\begin{equation}
\label{eq.19}
    \mathbf{H} = UV^T.
\end{equation}

\textit{3) Subproblem of $\bm{\omega}$}

With $\mathbf{H}$ and $\mathbf{G}_v$ fixed, Eq. (\ref{eq.14}) can be simply written as:
\begin{equation}
\label{eq.20}
	\begin{aligned}
		\underset{w_v}\max&\quad \sum_{v=1}^{V}\omega_{v}^{2}d_v  \\
		s.t.&\quad\bm{\omega}\mathbf{1}=1
	\end{aligned}
\end{equation}
where $d_v = \left \| \mathbf{K}_{v}-\mathbf{G}_{v} \mathbf{H}\right \|_{F}^{2}+\alpha \left \| \mathbf{G}_{v}-\mathbf{H}^T \right \|_{F}^{2}$, the problem can be solved by the method of Lagrange multipliers to obtain a closed-form solution:
\begin{equation}
\label{eq.21}
	w_v = \frac{1}{d_v\sum_{v=1}^{V}\frac{1}{d_v}  }. 
\end{equation}

\subsection{Initializaton}
In kernel-based multi-view initialization, eigenvector decomposition of the initial kernel matrix is typically performed using KKM. Beyond focusing on the kernel matrix, we also incorporate global relationships derived from the kernel matrix. When a sample has a high cumulative similarity to all other samples, it indicates that the sample is important. Thus, the initialization is obtained simply by solving the following formula:
\begin{equation}
\label{eq.22}
   	\underset{\mathbf{G}_v}\max \quad tr\left[\mathbf{G}_v^{T}\left(\mathbf{D}_v+\mathbf{K}_v\right)\mathbf{G}_v\right]
\end{equation}
where $\left(\mathbf{D}_v\right)_{ij} = \left\{\begin{matrix} 
   	\mathbf{A}_i^v,\quad i>j \\  
   	\mathbf{A}_j^v,\quad i<j 
   \end{matrix}\right.$ and $\mathbf{A}_i^v$ is the row sum of the kernel matrix $\mathbf{K}_v$ in the $i$-th row for the $v$-th view, representing the sum of similarities between the $i$-th sample and all other samples. Since $ \mathbf{D}_v + \mathbf{K}_v $ is still symmetric, we obtain  $\mathbf{G}_v$ by performing eigenvector decomposition on this matrix. Additionally, the initial matrix $\mathbf{H}$  is computed by averaging  $\mathbf{G}_v$  across views, followed by SVD decomposition.

In the optimization step, $\bm{\omega}$ is initialized as a vector where all elements are $\frac{1}{V}$.

Regarding the comparative algorithms, we directly download the publicly available MATLAB code and adjust the parameters according to the original articles. For algorithms involving k-means, we randomly repeat the process 50 times to determine the cluster centers. The only parameter $\alpha$ in our algorithm MVC-UMKLMF is determined using grid search within the range $\left [ 2^0,2^1,..,2^{9} \right ]$. The final clustering performance is evaluated using four widely used metrics: ACC (Accuracy), NMI (Normalized Mutual Information), Purity, and ARI (Adjusted Rand Index).

The entire process of alternating optimization is shown in Algorithm \ref{alg}.

\begin{algorithm}[ht]
	\caption{\textbf{MVC-UMKLMF}}
	\label{alg}
	\begin{algorithmic}[1]
		\STATE \algorithmicinput \hspace{1mm} Base kernel matrices $\left \{ \mathbf{K}_v \right \}_{v=1}^{V}$, hyperparameter $\alpha$ and cluster number k.

		\STATE \algorithmicoutput \hspace{1mm} Cluster labels.
		
		\STATE Initialize $\left \{ \mathbf{G}_v \right \}_{v=1}^{V}$, $\bm{\omega}$, and $\mathbf{H}$.
		
		\WHILE{not converged}
		\STATE Compute $\mathbf{G}_v$ by Eq. (\ref{eq.16});
		\STATE Compute $\mathbf{H}$ by Eq. (\ref{eq.19});
		\STATE Compute $w_v$ by Eq. (\ref{eq.21});
		\ENDWHILE
		
		\STATE Perform k-means on $\mathbf{H}$ to obtain final cluster labels.
	\end{algorithmic}
\end{algorithm}

\section{Theoretical Analysis} \label{sec4}
\subsection{Convergence}
The objective function denoted as $J_{\mathbf{H},\{\mathbf{G}_v\},\bm{\omega}}$ is monotonically decreasing when optimizing one variable while keeping the other variables fixed. Furthermore, it is evident that $J_{\mathbf{H},\{\mathbf{G}_v\},\bm{\omega}}$ is non-negative, monotonically decreasing, and has a lower bound; thus, the algorithm is theoretically convergent. We will further demonstrate the convergence in experiments in the following sections.
\subsection{Time Complexity}
The optimization part is divided into three subproblems. We will gradually analyze the time complexity of the three sub-problems in the optimization step. In the subproblem of $\mathbf{G}_v$, $\mathbf{G}_v$ is directly obtained from Eq. (\ref{eq.16}), and the time complexity of Eq. (\ref{eq.16}) is $O\left(kn^2\right)$. In the subproblem of $\mathbf{H}$, the complexity of calculating $\mathbf{A}$ is $O\left(Vkn^2\right)$, and the complexity of performing SVD decomposition on $\mathbf{A}$ is  $O\left(k^2n\right)$, and the complexity of computing $\mathbf{H}$ is $O\left(nk\right)$, so the computational complexity of this subproblem is $O\left(kn^2\right)$. In the subproblem of $\bm{\omega}$, the time complexity mainly focuses on calculating $d_v$, and the total time complexity is $O(V(kn^2+n^2))$. Since $V$ and $k$ are relatively small compared to $n$, the time complexity can be considered as $O\left(n^2\right)$.

Although the complexity of our algorithm still does not achieve a linear relationship with the number of samples, it has made progress compared to traditional multi-kernel and graph-based methods, which will be inspiring to existing approaches.

\begin{table}[!t]
    \centering
    \caption{Datasets Used in the Experiments\label{table1}}
    \begin{tabular}{|c|c|c|c|c|}
        \hline
        Datasets           & Types                    & Samples & Clusters & Views \\ \hline
        Texas              & Text                     & 187     & 5        & 2     \\
        Cornell            & Text                     & 195     & 5        & 2     \\
        MSRA               & Image                    & 210     & 7        & 5     \\
        Caltech101-7       & Image                    & 441     & 7        & 6     \\
        BBCSports          & Text                     & 544     & 5        & 2     \\
        BBC                & Text                     & 685     & 5        & 4     \\
        ProteinFold        & Bioinformatics   & 694     & 27       & 12    \\
        Caltech101-20      & Image                    & 2386    & 20       & 6     \\
        CCV                & Video                    & 6773    & 20       & 3     \\
        Caltech101-all     & Image                    & 9144    & 102      & 6\\
     
        \hline
    \end{tabular}
\end{table}

\begin{figure*}[!ht]
    \centering
    \subfloat[]{
        \includegraphics[width=0.23\textwidth]{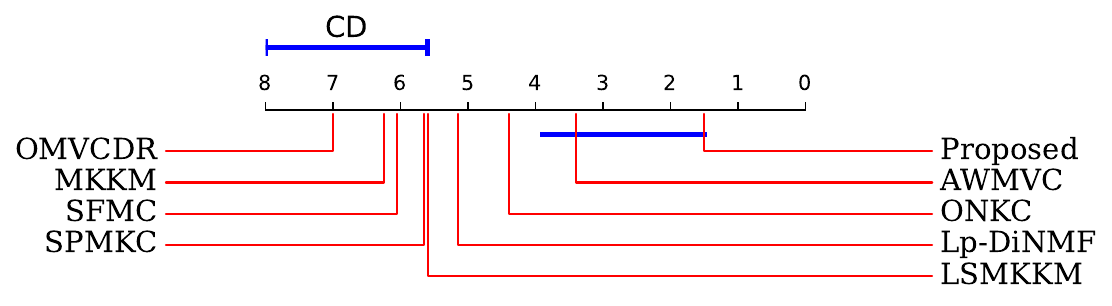} 
        \label{fig2a}
    }
    \hfil
    \subfloat[]{
        \includegraphics[width=0.23\textwidth]{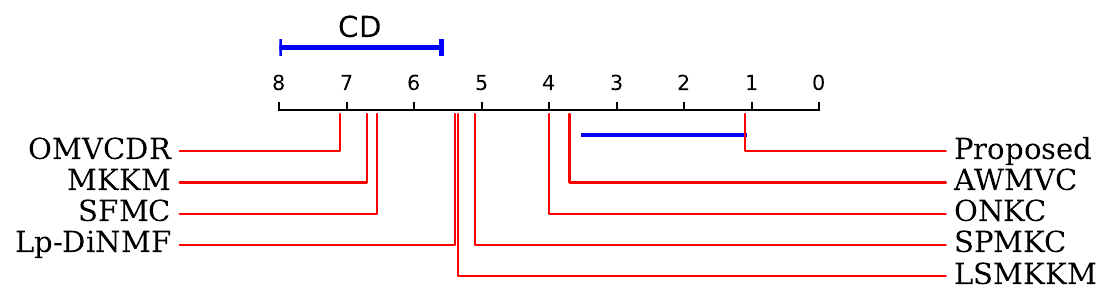} 
        \label{fig2b}
    }
    \hfil
    \subfloat[]{
        \includegraphics[width=0.23\textwidth]{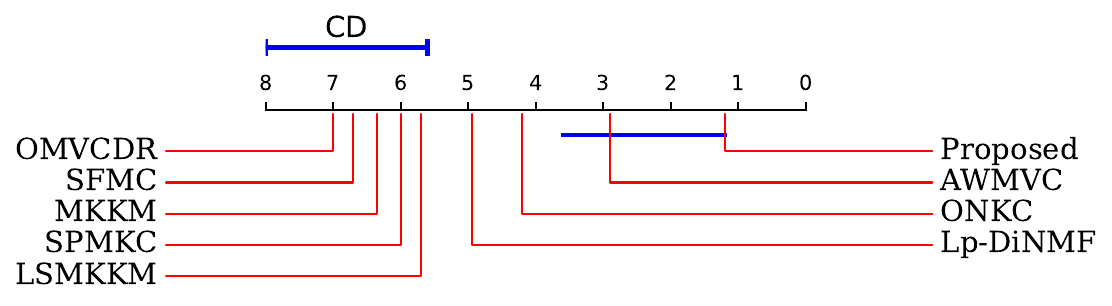} 
        \label{fig2c}
    }
    \hfil
    \subfloat[]{
        \includegraphics[width=0.23\textwidth]{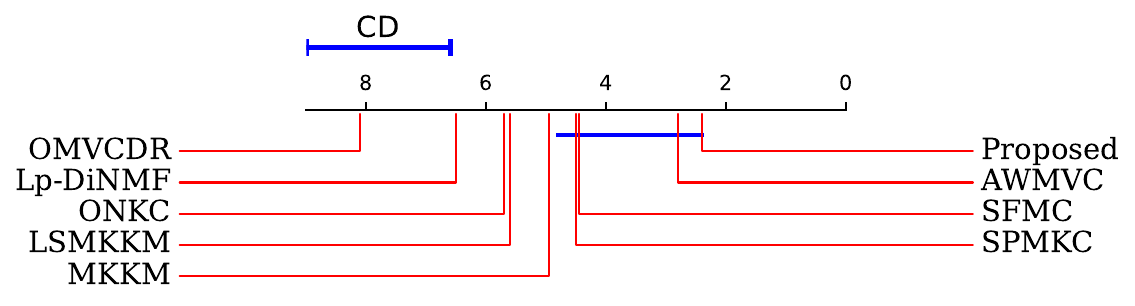} 
        \label{fig2d}
    }
   
    \caption{Comparison of nine clustering algorithms using Nemenyi’s test. (a) Statistical Results on the ACC Metric. (b) Statistical Results on the NMI Metric. (c) Statistical Results on the Purity Metric. (d) Statistical Results on the ARI Metric.}
    \label{fig2}
\end{figure*}

\section{Experiment} \label{sec5}
\subsection{Experiment Setting}
In this section, we will demonstrate the effectiveness of our proposed method, MVC-UMKLMF, by selecting 10 representative multi-view datasets. The evaluation of MVC-UMKLMF will be conducted in terms of clustering performance, statistical significance test of algorithms, experimental convergence, visualization and parameter sensitivity.

\begin{figure*}[!htbp]
	\centering
	\subfloat[]{
		\includegraphics[width=0.18\textwidth]{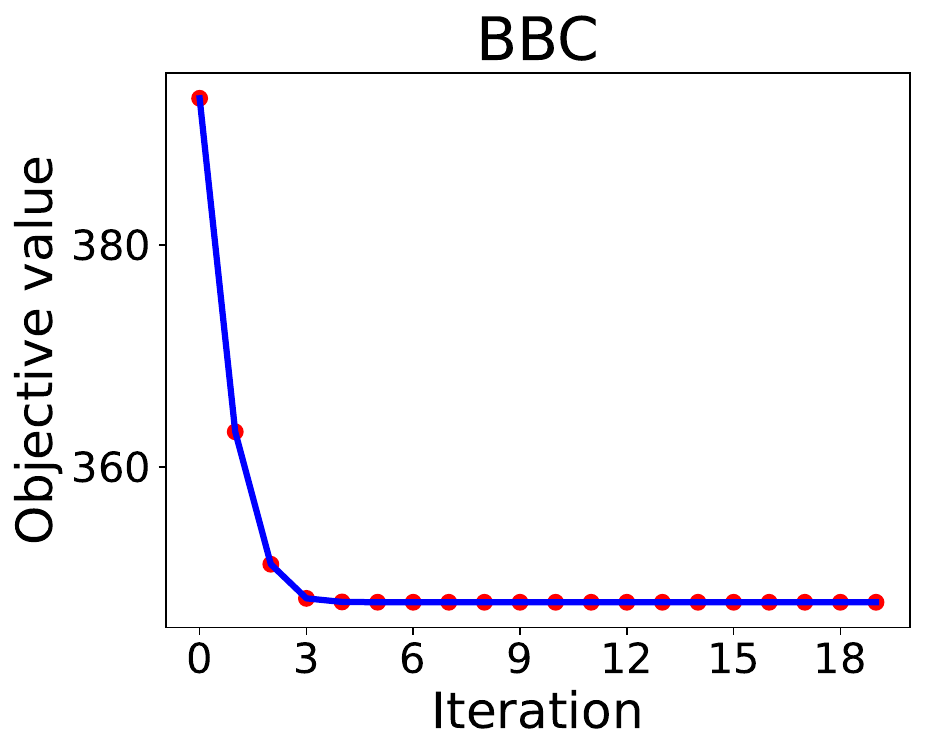} 
		\label{fig3a}
	}
	\hfil
	\subfloat[]{
		\includegraphics[width=0.18\textwidth]{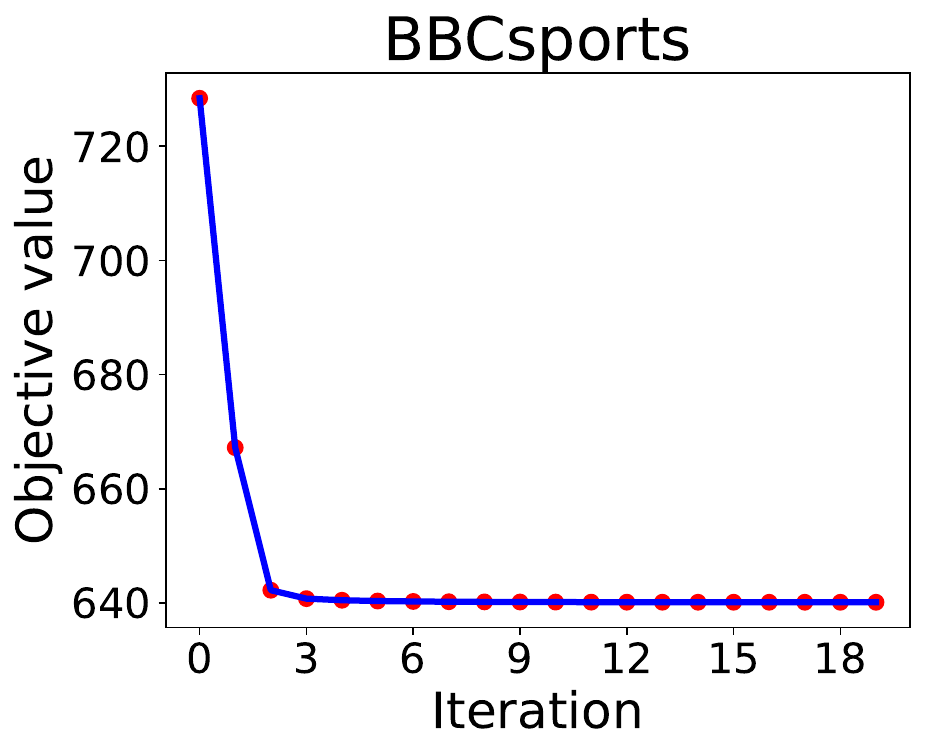}
		\label{fig3b}
	}
	\hfil
	\subfloat[]{
		\includegraphics[width=0.18\textwidth]{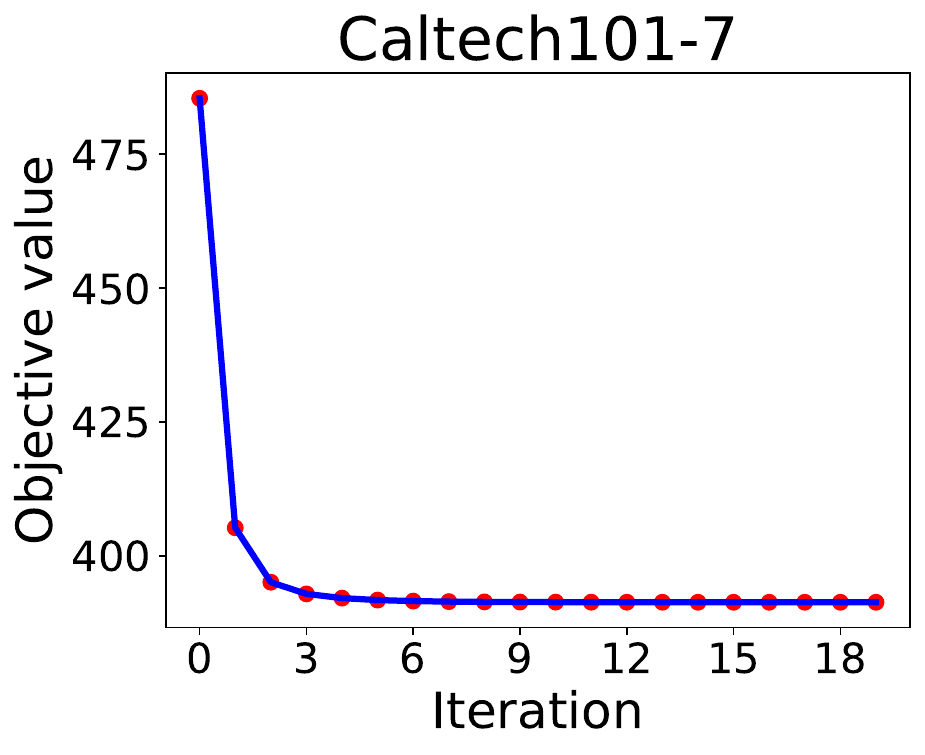}
		\label{fig3c}
	}
	\hfil
	\subfloat[]{
		\includegraphics[width=0.18\textwidth]{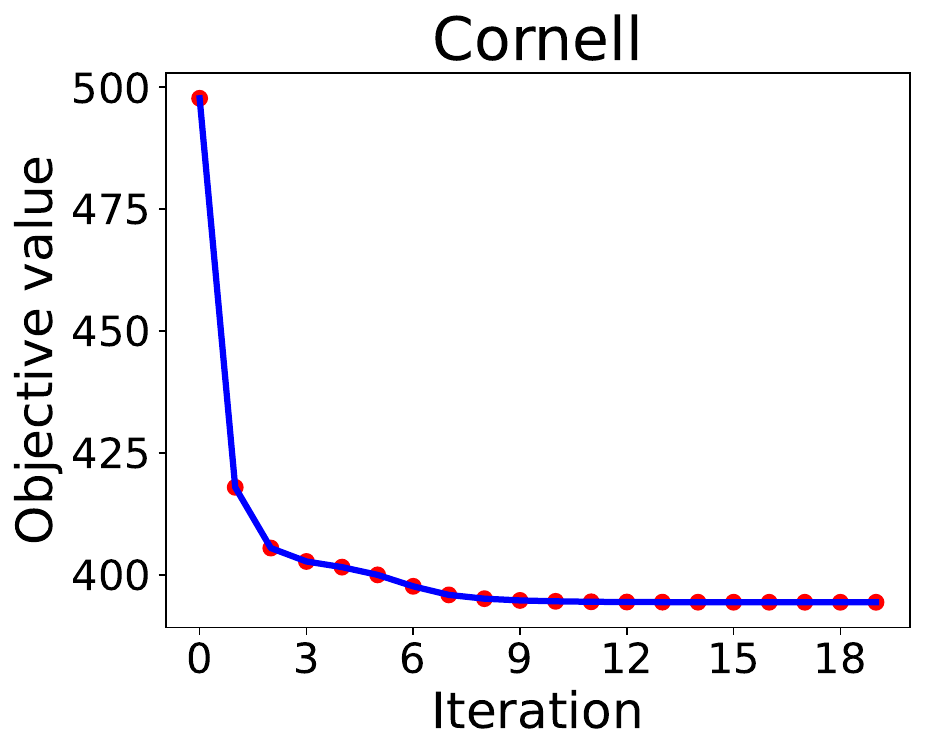}
		\label{fig3d}
	}
	\hfil
	\subfloat[]{
		\includegraphics[width=0.18\textwidth]{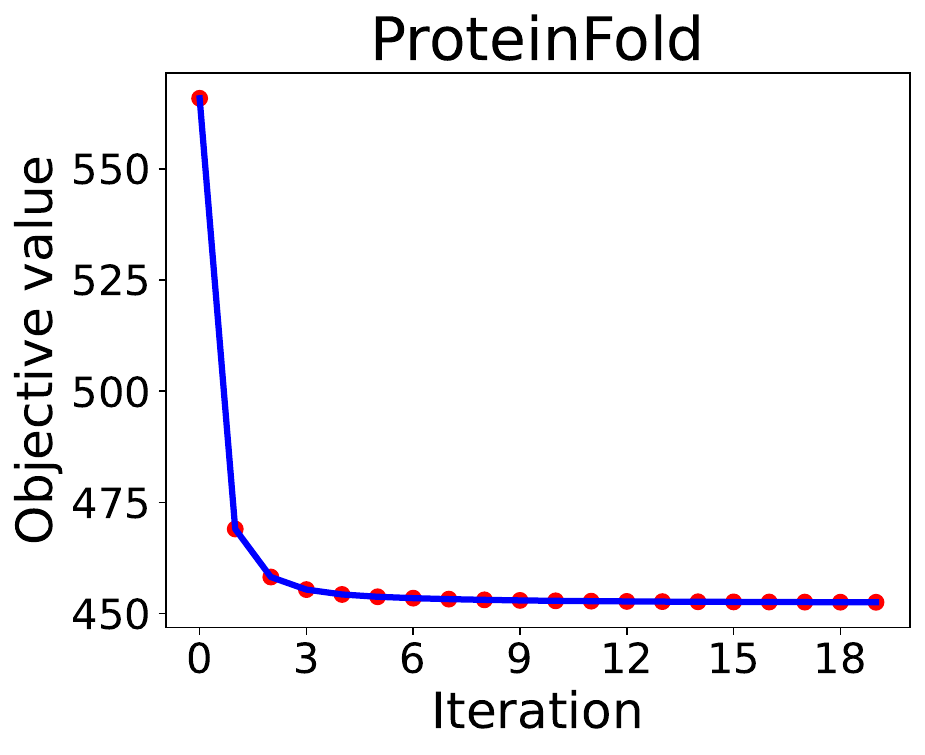}
		\label{fig3e}
	}
	\caption{Convergence of the proposed method on six datasets. (a) Convergence on BBC. (b) Convergence on BBCsports. (c) Convergence on Caltech101-7. (d) Convergence on Cornell. (e) Convergence on ProteinFold.}
	\label{fig3}
\end{figure*}

\begin{figure}[!t]
	\centering
	\subfloat[]{%
		\includegraphics[width=0.22\textwidth]{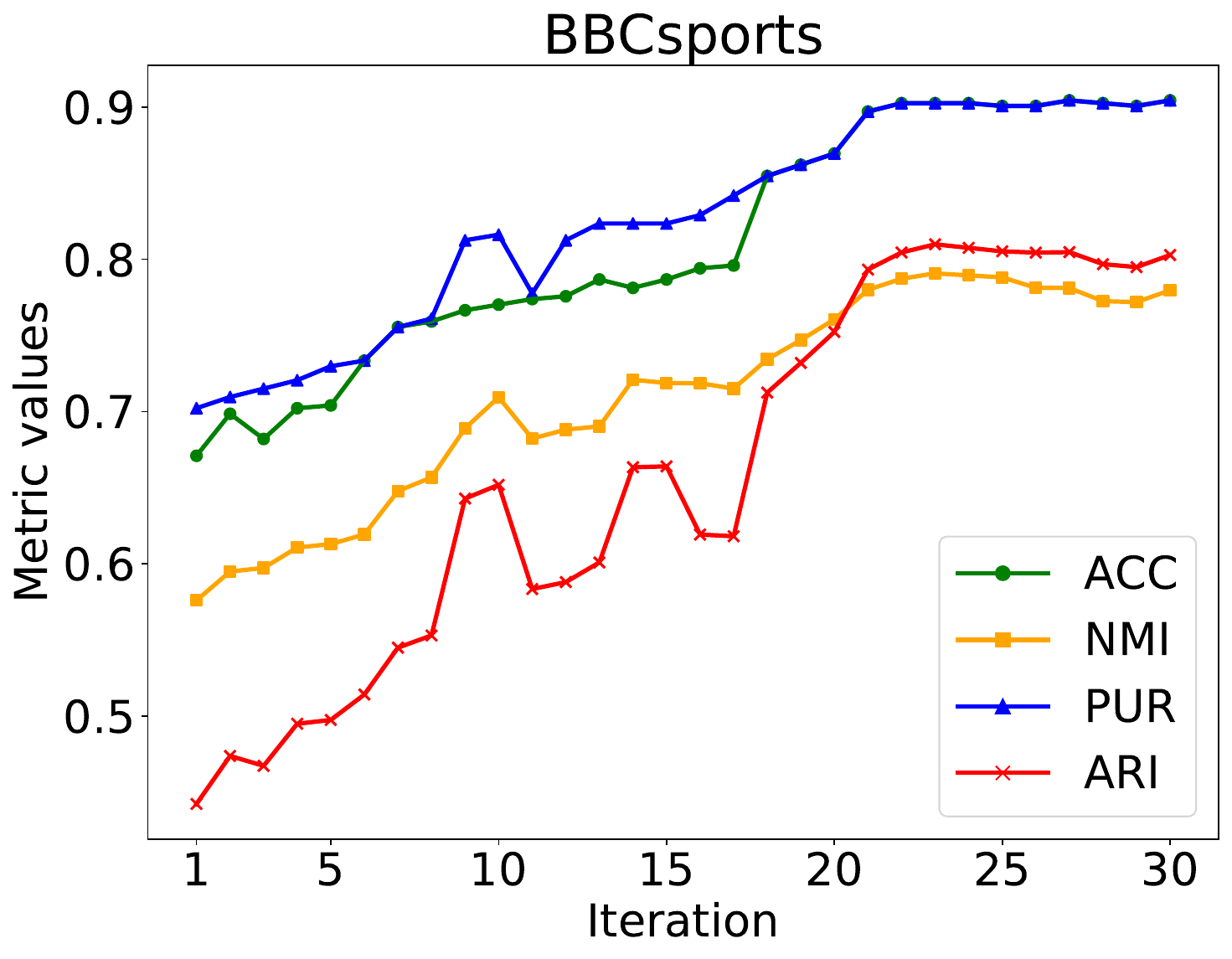}%
		\label{fig4a}
	}
    \hfill
	\subfloat[]{%
		\includegraphics[width=0.22\textwidth]{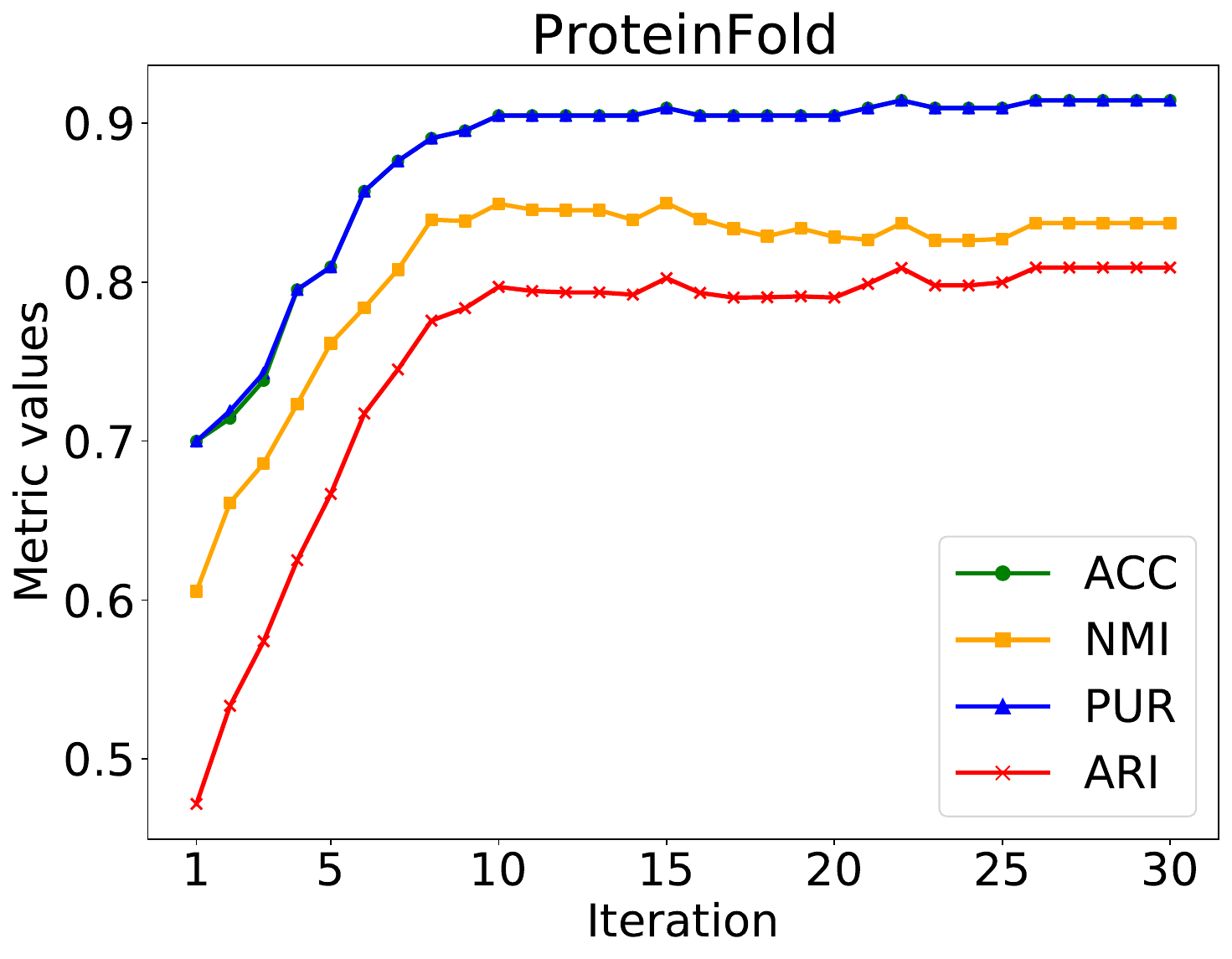}%
		\label{fig4b}
	}
	\caption{Clustering Performance over Iteration on Two Datasets. (a) BBCsports. (b) ProteinFold.}
	\label{fig4}
\end{figure}

\subsubsection{Datasets}The datasets we have selected include texas\footnotemark[1]\footnotetext[1]{{\url{http://www.cs.cmu.edu/afs/cs.cmu.edu/project/theo-20/www/data/}}}, Cornell\footnotemark[1], MSRA\cite{winnLOCUSLearningObject2005}, Caltech101-7\footnotemark[2]\footnotetext[2]{{\url{http://www.vision.caltech.edu/Image_Datasets/Caltech101/}}}, BBCsports\footnotemark[3]\footnotetext[3]{{\url{http://mlg.ucd.ie/datasets/bbc.html}}}
, BBC, proteinfold\footnotemark[4]\footnotetext[4]{{\url{mkl.ucsd.edu/dataset/protein-fold-prediction}}}, Caltech101-20,  CCV\footnotemark[5]\footnotetext[5]{{\url{https://www.ee.columbia.edu/ln/dvmm/CCV/}}} and Caltech101-all. Detailed information about the datasets can be found in Table \ref{table1}. All the precomputed base kernels within the datasets are publicly available on websites.
\subsubsection{Comparative Algorithms} We have selected 8 comparative algorithms, covering multi-kernel learning, matrix factorization, and graph-based approaches. The basic information about these algorithms is as follows.\\
\begin{itemize}
\item{MKKM (Huang et al. 2012) \cite{hsin-chienhuangMultipleKernelFuzzy2012}: The optimal kernel is considered to be a linear combination of base kernels, and the algorithm simultaneously optimizes the base kernel coefficients and the low-dimensional embedding matrix $\mathbf{H}$.}
\item{ONKC (Liu et al. 2017) \cite{liuOptimalNeighborhoodKernel2017}: It learns the optimal kernel from the neighborhood of linear kernels, breaking away from the original limitation that the optimal kernel must be a linear kernel, thus learning the optimal kernel matrix within a broader range.}
\item{LP-DiNMF (Wang et al. 2018) \cite{wangDiverseNonNegativeMatrix2018}: It considers the specificity between views, eliminates redundancy between views, and incorporates local supervisory information of data points to learn low-dimensional representations of each view, eventually fusing them to learn a consensus representation.}
\item{SPMKC (Ren et al. 2021) \cite{rensimultaneousGlobal2021}:It enhances the clustering performance of multi-kernel learning by simultaneously preserving the global and local graph structures of the data.}
\item{LSMKKM (Liu et al. 2021) \cite{liuLocalizedSimpleMultiple2021}: It achieves a local version of simpleMKKM by aligning local kernels.}
\item{SFMC (Li et al. 2022) \cite{liMultiviewClusteringScalable2022}: It presents a multi-view graph fusion framework that exploits the joint graph across multiple views via a self-supervised weighting manners without extra smoothness parameters.}
\item{AWMVC (Wan et al. 2023) \cite{wanAutoweightedMultiviewClustering2023}: It removes the non-negativity constraint and obtains coefficient matrices of different dimensions, integrating the coefficient matrices into a consensus matrix for a richer data representation.}
\item{OMVCDR (Wan et al. 2024) \cite{wanOneStepMultiViewClustering2024a}: It improves the way of integrating consensus on the basis of AWMVC and obtains clustering results in one step.}
\end{itemize}

\begin{table*}[ht]
	\caption{Results of Experiment\\The Best Results are Highlighted in Red, and the Second-best Results are Highlighted in Blue, '-' Indicates that the Algotithm cannot be Executed Smoothly Due to the Out-of-Memory or Other Reasons}
	\label{table2}
	\centering
	{\small
		\resizebox{\linewidth}{!}{
			\begin{tabular}{ccccccccccc}
				\toprule
				Datasets&Metrics&MKKM&ONKC&LP-DiNMF&SPMKC&LSMKKM&SFMC&AWMVC&OMVCDR&Proposed\\
				\midrule
				\multirow{4}{*}{Texas}&ACC&0.5615&0.5455&\textcolor{red}{0.6684}&0.5241&0.3666&0.5561&\textcolor{blue}{0.6257}&-&0.5722\\
				&NMI&0.0459&0.3140&\textcolor{blue}{0.3487}&0.1701&0.1764&0.0593&0.3186&-&\textcolor{red}{0.3667}\\
				&Purity&0.5722&0.7112&0.7326&0.5668&0.6267&0.5668&\textcolor{red}{0.7433}&-&\textcolor{red}{0.7380}\\
				&ARI&\textcolor{blue}{0.5536}&0.2360&0.3907&0.5119&0.1071&0.5444&\textcolor{red}{0.5863}&-&0.2690\\
				\midrule
				\multirow{4}{*}{Cornell}&ACC&0.4462&0.5179&0.5436&0.3846&0.5285&0.4462&\textcolor{blue}{0.5846}&0.4359&\textcolor{red}{0.6308}\\
				&NMI&0.0387&0.3588&\textcolor{blue}{0.3723}&0.0817&0.3025&0.0619&0.3136&0.1591&\textcolor{red}{0.4410}\\
				&Purity&0.4513&0.6359&\textcolor{blue}{0.6667}&0.4718&0.6413&0.4513&0.6462&0.5179&\textcolor{red}{0.6974}\\
				&ARI&0.4212&0.2833&0.2618&0.3753&0.2362&\textcolor{blue}{0.4299}&\textcolor{red}{0.4919}&0.1161&0.3994\\
				\midrule
				\multirow{4}{*}{MSRA}&ACC&0.5524&0.8000&0.6571&0.7762&0.7171&0.8095&\textcolor{blue}{0.8429}&0.6238&\textcolor{red}{0.9048}\\
				&NMI&0.4854&0.6747&0.5738&0.6771&0.6729&\textcolor{blue}{0.7412}&0.7270&0.5427&\textcolor{red}{0.8459}\\
				&Purity&0.5619&0.8000&0.6857&0.7762&0.7693&0.8095&\textcolor{blue}{0.8429}&0.6238&\textcolor{red}{0.9048}\\
				&ARI&0.4514&0.6145&0.4444&0.6398&0.5975&0.7136&\textcolor{blue}{0.7221}&0.4371&\textcolor{red}{0.7951}\\
				\midrule
				\multirow{4}{*}{Caltech101-7}&ACC&0.4966&0.5828&0.4098&0.6712&\textcolor{blue}{0.7071}&0.6526&0.4518&0.5529&\textcolor{red}{0.8095}\\
				&NMI&0.4818&0.5437&0.2680&\textcolor{blue}{0.6278}&0.5629&0.5629&0.5304&0.3292&\textcolor{red}{0.7749}\\
				&Purity&0.6168&0.6689&0.7822&0.7120&0.7989&\textcolor{blue}{0.8528}&0.8433&0.8107&\textcolor{red}{0.8639}\\
				&ARI&0.4502&0.4202&0.2294&0.5994&\textcolor{blue}{0.6573}&0.6409&0.4961&0.3551&\textcolor{red}{0.7882}\\
				\midrule
				\multirow{4}{*}{BBCsports}&ACC&0.6048&0.5938&\textcolor{red}{0.9118}&0.8658&0.5028&0.3603&0.6397&0.4357&\textcolor{blue}{0.9044}\\
				&NMI&0.4420&0.4200&\textcolor{blue}{0.7726}&0.7495&0.2671&0.0239&0.4820&0.1180&\textcolor{red}{0.7783}\\
				&Purity&0.7335&0.7261&\textcolor{red}{0.9118}&0.8658&0.5406&0.3640&0.6893&0.4375&\textcolor{blue}{0.9044}\\
				&ARI&0.5329&0.3867&0.7838&\textcolor{red}{0.8398}&0.2456&0.3870&0.5234&0.0973&\textcolor{blue}{0.8027}\\
				\midrule
				\multirow{4}{*}{BBC}&ACC&0.5693&0.5956&\textcolor{blue}{0.8146}&\textcolor{blue}{0.8146}&0.4923&0.3328&0.6555&0.4423&\textcolor{red}{0.8248}\\
				&NMI&0.3120&0.3366&\textcolor{blue}{0.5992}&0.5878&0.2225&0.0277&0.4133&0.2080&\textcolor{red}{0.7086}\\
				&Purity&0.5737&0.6000&\textcolor{blue}{0.8146}&\textcolor{blue}{0.8146}&0.5264&0.3387&0.6555&0.5255&\textcolor{red}{0.8248}\\
				&ARI&0.4192&0.2927&0.6447&\textcolor{red}{0.7235}&0.2443&0.3806&0.5138&0.1785&\textcolor{blue}{0.6692}\\
				\midrule
				\multirow{4}{*}{ProteinFold}&ACC&0.2594&\textcolor{blue}{0.3559}&0.1153&0.1844&0.3082&0.1945&0.3429&0.2305&\textcolor{red}{0.4784}\\
				&NMI&0.3173&\textcolor{blue}{0.4448}&0.0458&0.2260&0.4292&0.2012&0.4293&0.2843&\textcolor{red}{0.6155}\\
				&Purity&0.2997&0.4207&0.1484&0.2320&0.3875&0.2147&\textcolor{blue}{0.4092}&0.2882&\textcolor{red}{0.5663}\\
				&ARI&0.1587&0.1736&0.0046&0.1169&0.1695&0.1017&\textcolor{blue}{0.1992}&0.0852&\textcolor{red}{0.3289}\\
				\midrule
				\multirow{4}{*}{Caltech101-20}&ACC&0.3013&0.3713&0.3474&\textcolor{red}{0.6312}&0.3148&\textcolor{blue}{0.6018}&0.4845&0.4216&0.4992\\
				&NMI&0.3918&0.5437&0.3783&\textcolor{red}{0.6804}&0.3418&0.5654&0.6186&0.4228&\textcolor{blue}{0.6307}\\
				&Purity&0.6370&0.7469&0.6303&0.7741&0.4810&0.7008&0.7795&0.5905&\textcolor{red}{0.8286}\\
				&ARI&0.2670&0.2851&0.2212&\textcolor{blue}{0.6175}&\textcolor{red}{0.7086}&0.4333&0.4127&0.3309&0.3543\\
				\midrule
				\multirow{4}{*}{CCV}&ACC&0.1801&\textcolor{blue}{0.2538}&0.1045&-&0.1739&0.1155&0.1897&0.1506&\textcolor{red}{0.3185}\\
				&NMI&0.1487&\textcolor{blue}{0.2020}&0.0028&-&0.1541&0.0360&0.1404&0.1275&\textcolor{red}{0.4276}\\
				&Purity&0.2178&\textcolor{blue}{0.2817}&0.1072&-&0.2168&0.1202&0.2226&0.1849&\textcolor{red}{0.3775}\\
				&ARI&0.1085&0.0905&0.0007&-&0.0574&0.1085&\textcolor{blue}{0.1104}&0.0433&\textcolor{red}{0.1937}\\
				\midrule
				\multirow{4}{*}{Caltech101-all}&ACC&0.1534&0.2629&0.2083&-&0.1964&0.1428&\textcolor{blue}{0.2940}&0.1601&\textcolor{red}{0.4891}\\
				&NMI&0.3351&0.4814&0.4210&-&0.4239&0.2093&\textcolor{blue}{0.5196}&0.3499&\textcolor{red}{0.7325}\\
				&Purity&0.3262&0.4857&0.4233&-&0.4168&0.2022&\textcolor{blue}{0.5016}&0.3044&\textcolor{red}{0.7422}\\
				&ARI&0.1105&0.1908&0.1529&-&0.1473&0.0481&\textcolor{blue}{0.2682}&0.1068&\textcolor{red}{0.3253}\\
				\bottomrule
			\end{tabular}
	}}
\end{table*}

\subsection{Experimental Results} 
We presented the performance of the proposed method and nine comparison algorithms on four clustering metrics: ACC, NMI, purity, and ARI across ten datasets, as shown in Table \ref{table2}. For each dataset and metric, the best result is highlighted in red, and the second-best result is highlighted in blue. The results in Table \ref{table2} are summarized as follows.

The proposed method ranks in the top two overall, ranking first on most datasets. On the MSRA dataset, the proposed method exceeds the second-ranked method by 6, 10, 6, and 7 percentage points in ACC, NMI, purity, and ARI, respectively. On the proteinfold dataset, it exceeds the second-ranked method by 12, 17, 16, and 13 percentage points, and on the CCV dataset, it exceeds by 13, 27, 15, and 8 percentage points. Particularly, the performance on the NMI metric is outstanding.

Among matrix factorization-based methods, LP-DiNMF and AWMVC are highly competitive, ranking second on many datasets. LP-DiNMF aims to ensure that each view has a distinct embedding, with diversity as its main focus. In contrast, our proposed method learns embeddings for each view spontaneously without enforcing distinctness, as we believe that forced diversity between views may not be applicable to all datasets. We attribute the strong performance of AWMVC to its removal of the non-negativity constraint, which is consistent with our approach. AWMVC learns multi-dimensional embeddings for each view, recognizing the ambiguity in the structure of a single view. We share this consideration but adopted a different approach by discarding the orthogonality constraint for individual views.

The graph-based method SPMKC performs well on certain datasets, especially on the Cal101-20 dataset, the graph-based method SFMC also achieved leading results. We infer that the affinity graph learned through the kernel matrix has an adequately accurate number of blocks, leading to a clearer learned clustering structure. In contrast, kernel-based methods are limited to the space of linear kernels, resulting in less outstanding performance.

\subsection{Statistical Significance Test of Algorithms}
In order to provide a clearer explanation of the experimental results, we will conduct some statistical experiments in this section. The Friedman's test is a non-parametric statistical test used to compare differences among multiple related samples or groups under repeated measures. Here, the Friedman's test can be used to indicate whether there are significant differences in the performance of all algorithms across different datasets. However, its test statistic often underestimates the differences between groups\cite{friedmanUseRanksAvoid1937}. Therefore, we use the Iman-Davenport extended Friedman's test, which transforms the $\chi^2$ distribution into an F distribution, with a confidence level set at $\alpha = 0.05$.

\begin{figure*}[!htbp]
    \centering
    \subfloat[]{
        \includegraphics[width=0.15\textwidth]{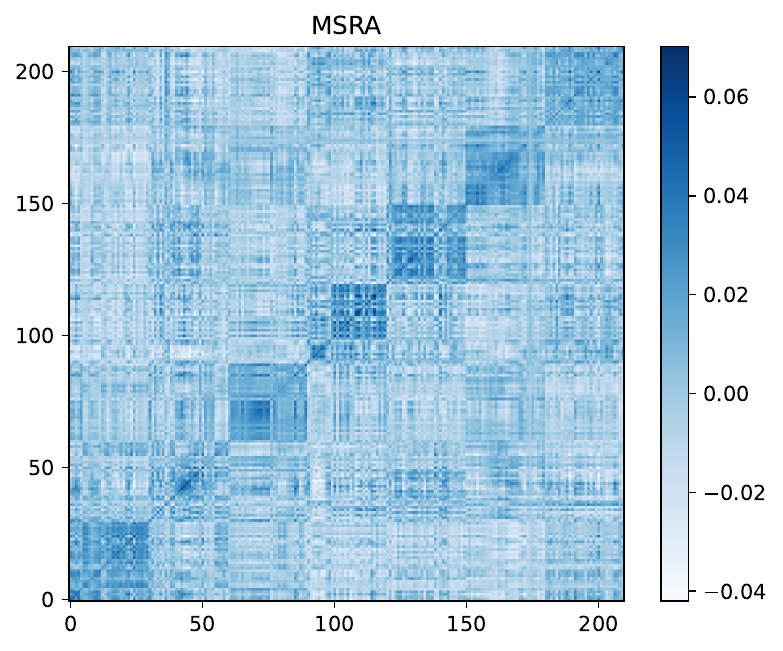} 
        \label{fig5a}
    }
    \hfil
    \subfloat[]{
        \includegraphics[width=0.15\textwidth]{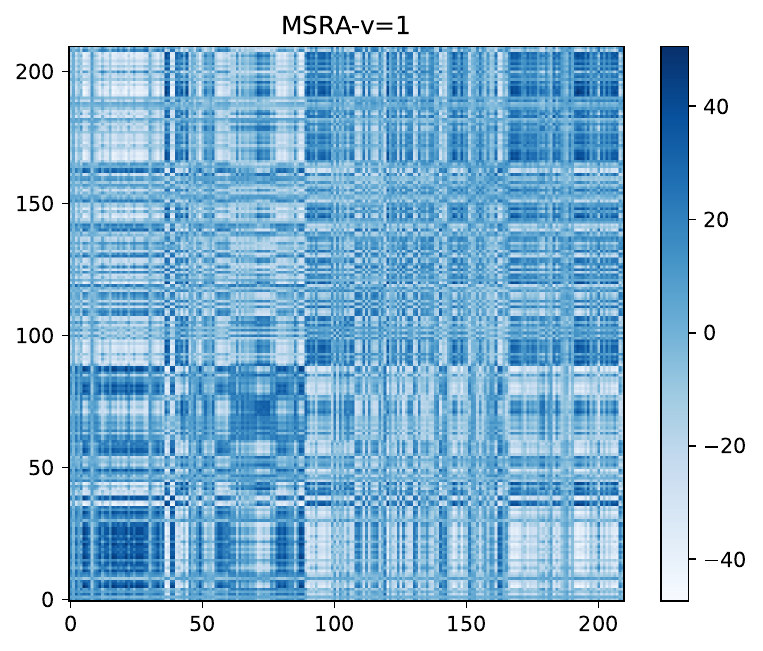} 
        \label{fig5b}
    }
    \hfil
    \subfloat[]{
        \includegraphics[width=0.15\textwidth]{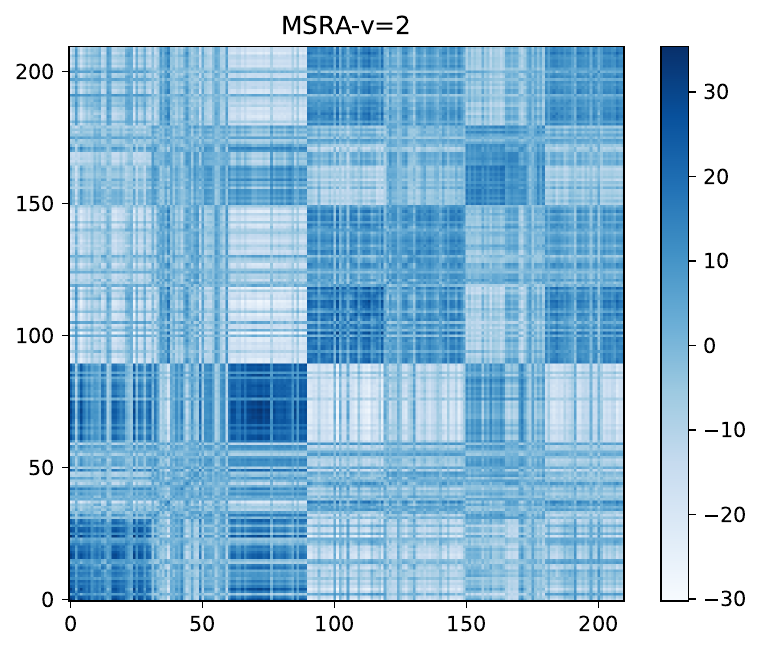} 
        \label{fig5c}
    }
    \hfil
    \subfloat[]{
        \includegraphics[width=0.15\textwidth]{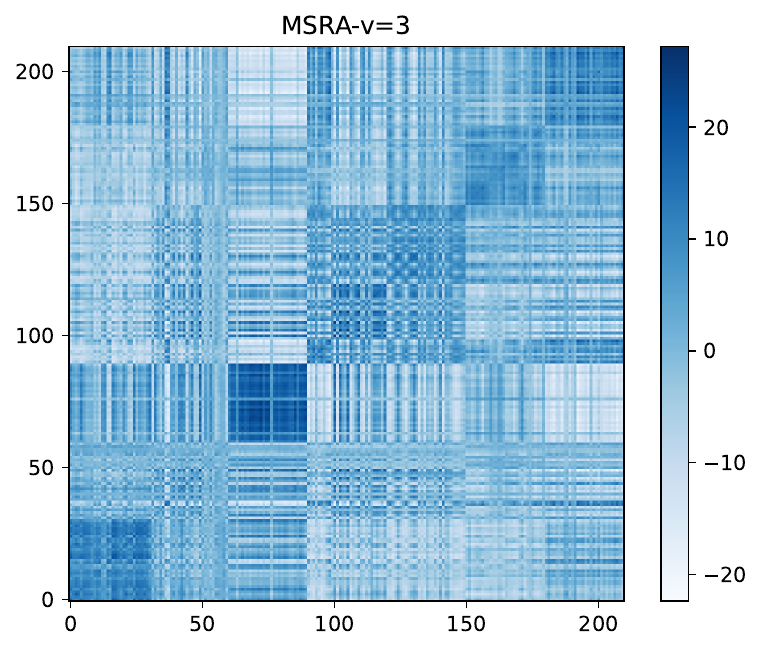} 
        \label{fig5d}
    }
    \hfil
    \subfloat[]{
        \includegraphics[width=0.15\textwidth]{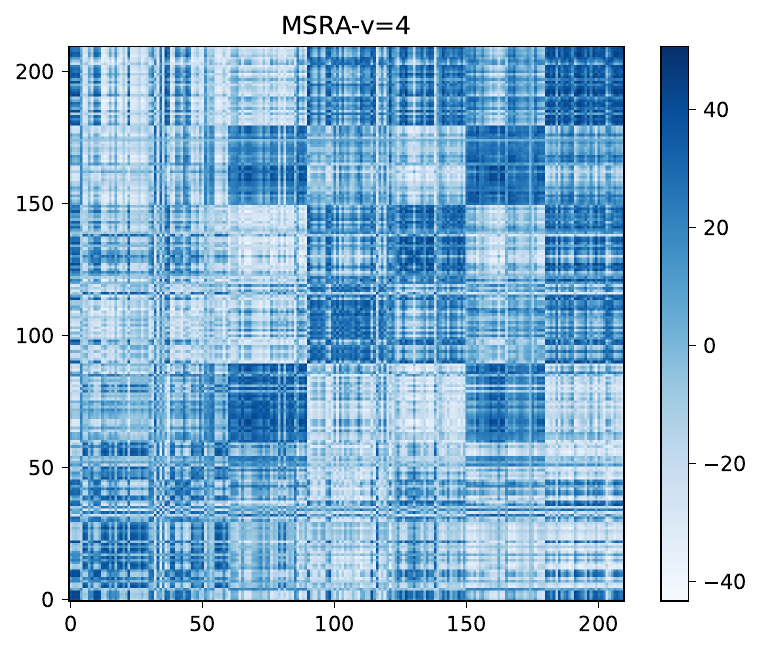} 
        \label{fig5e}
    }
    \hfil
    \subfloat[]{
        \includegraphics[width=0.15\textwidth]{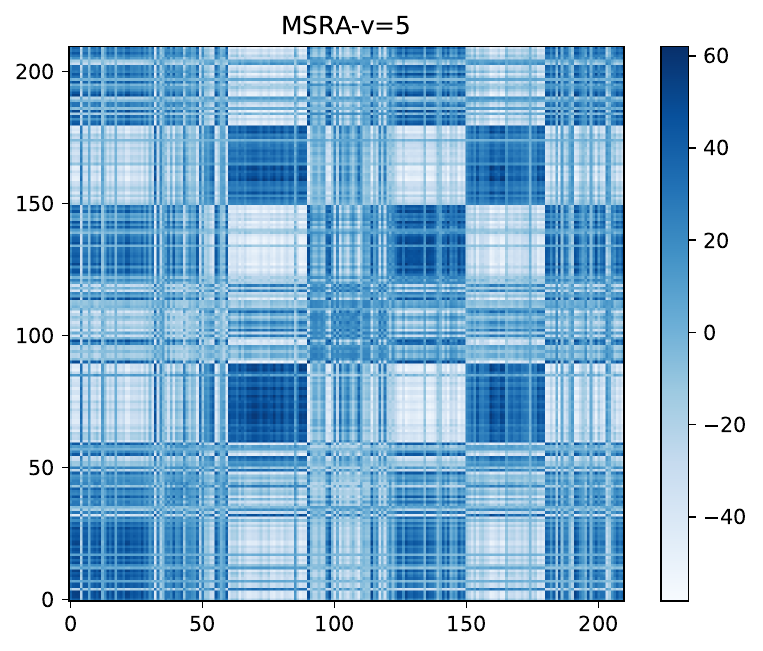} 
        \label{fig5f}
    }
    \caption{Visualization of different views in MSRA. (a) Visualization of $\mathbf{H}\mathbf{H}^T$. (b) Visualization of $\mathbf{G}_1\mathbf{G}_1^T$. (c) Visualization of $\mathbf{G}_2\mathbf{G}_2^T$. (d) Visualization of $\mathbf{G}_3\mathbf{G}_3^T$. (e) Visualization of $\mathbf{G}_4\mathbf{G}_4^T$. (f) Visualization of $\mathbf{G}_5\mathbf{G}_5^T$.}
    \label{fig5}
\end{figure*}

Table \ref{table3} presents some numerical results from the statistical tests. The $p$ values obtained from the four metrics are much smaller than the significance level $\alpha$ . Therefore, it can be concluded that there are significant differences among the 9 algorithms.

\begin{figure}[!htbp]
    \centering
    \subfloat[]{
        \includegraphics[width=0.14\textwidth]{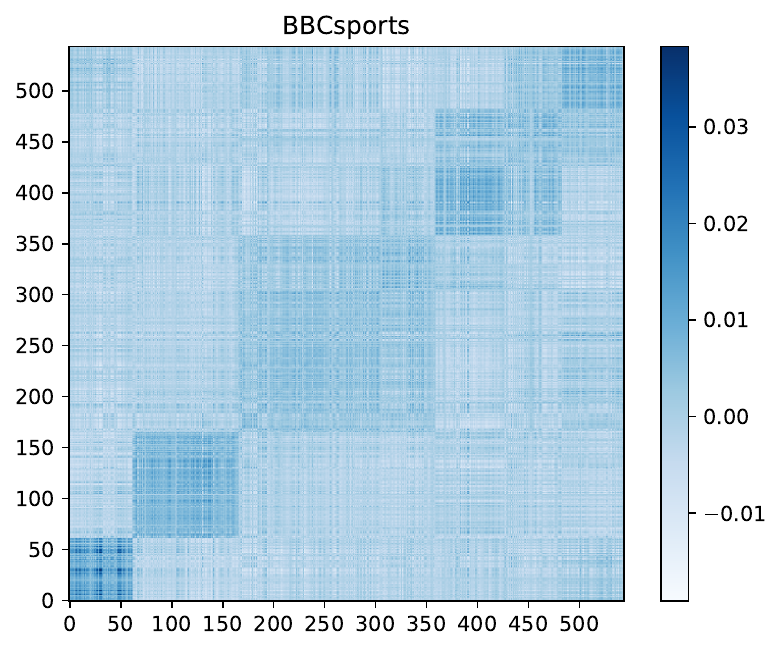} 
        \label{fig6a}
    }
    \hfil
    \subfloat[]{
        \includegraphics[width=0.14\textwidth]{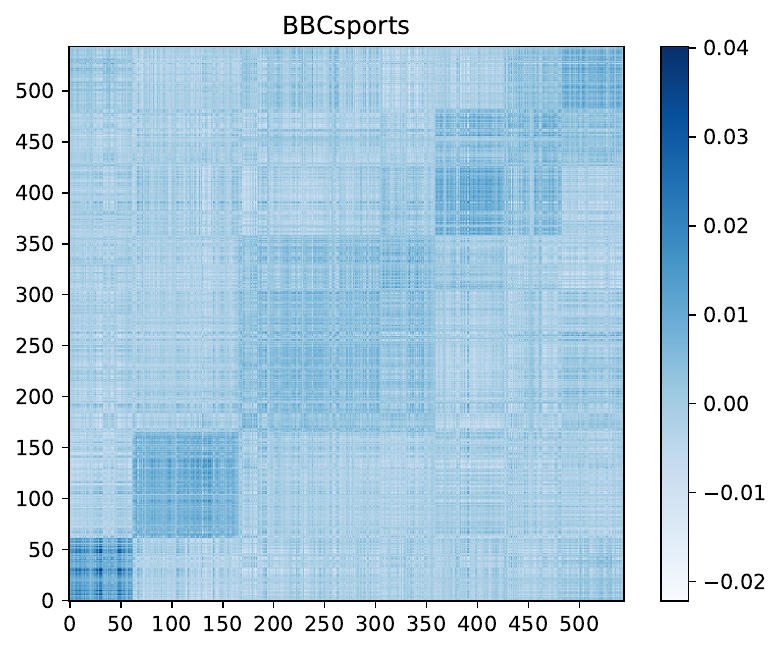} 
        \label{fig6b}
    }
    \hfil
    \subfloat[]{
        \includegraphics[width=0.14\textwidth]{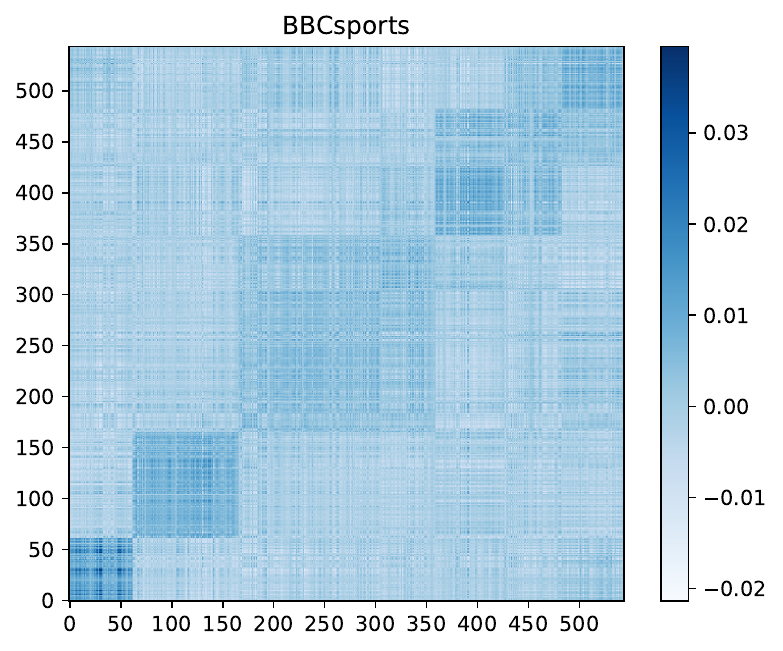} 
        \label{fig6c}
    }
    \caption{Visualization of different views in BBCsports. (a) Visualization of $\mathbf{H}\mathbf{H}^T$. (b) Visualization of $\mathbf{G}_1\mathbf{G}_1^T$. (c) Visualization of $\mathbf{G}_2\mathbf{G}_{2}^{T}$. }
    \label{fig6}
\end{figure}

\begin{table}[!t]
\caption{Numerical Results of Statistical Significance Test\label{table3}}
\centering	
\begin{tabular}{|c|c|c|c|c|}
	\hline
		Metrics           & F                   & df1 & df2 &p    \\ 
		\hline
		ACC              & 5.4540                     & 8     & 72        &2.2051e-5     \\
		NMI            & 8.8188                     & 8     & 72        & 1.6875e-7     \\
		Purity               & 7.8198                    & 8     & 72        & 2.5230e-8     \\
		ARI       & 6.3693                    & 8     & 72        & 3.1401e-6     \\
		
		\hline
\end{tabular}
\end{table}

We continue to use Nemenyi's test to further determine the pairwise differences between the algorithms. We calculate the average rank of each algorithm across different datasets\cite{demsarStatisticalComparisonsClassifiers2006a}. If the difference in ranks is less than the critical difference (CD), it indicates that the statement of no significant difference between the two algorithms will be accepted with 95\% confidence. The calculation of CD is given by the Eq. (\ref{eq.23}).
\begin{equation}
\label{eq.23}
    \text{CD} = q_{\alpha} \sqrt{\frac{k(k + 1)}{6n}}
\end{equation}
where $k$ represents the number of algorithms and $n$ represents the number of datasets. $q_{\alpha}=1.96$, and the calculated CD is 2.4004. The results of the Nemenyi's test are shown in Figure \ref{fig2}. At a confidence level of 0.05, for NMI, our method shows significant differences compared to all other algorithms. However, for ARI, our method does not demonstrate outstanding performance. Nevertheless, our method still has the highest overall ranking, showing a leading advantage in other metrics. Furthermore, except for our method and AWMVC, there are no significant differences among the other methods, which indirectly indicates the effectiveness of the proposed method.
\subsection{Convergence and Evolution}
We selected five datasets to plot the objective function's variation with the number of iterations, as shown in Figure \ref{fig3}. The objective function for all six datasets decreases and converges within 10 iterations. Besides, Figure \ref{fig4} shows the changes in clustering metrics, including ACC, NMI, Purity, and ARI, with the number of iterations on two datasets. It can be observed that as H iterates, the metric results first show an increasing trend and then stabilize, demonstrating the effectiveness of our proposed algorithm.

\subsection{Visualization}
The final learned matrix $\mathbf{H}$ represents the low-dimensional embedding of the original data, and $\mathbf{H}\mathbf{H}^T$ can be expressed as a similarity matrix. In well-organized datasets, it should exhibit a clear block-diagonal structure. Similarly, $\mathbf{G}_v$  represents the embedding of a single view, and  $\mathbf{G}_v\mathbf{G}_v^T$ represents the similarity matrix for a single view. To visually demonstrate the effectiveness, we show the visualization results for the MSRA and BBCsports datasets in Figure \ref{fig5} and Figure \ref{fig6}. On the MSRA dataset, the block-diagonal structure of $\mathbf{H}\mathbf{H}^T$ is the clearest, with the number of blocks corresponding to the number of clusters. In contrast, the structure for other views appears more disorganized, indicating that the clustering structure within a single view is neither clear nor reliable, making it unreasonable to impose orthogonal constraints on it. On the BBCsports dataset, there is little difference between the single-view and consensus embeddings, suggesting that the information from a single view is sufficient, a phenomenon that exists in some datasets. Due to the varying characteristics of different datasets, it is not feasible to enforce a standard clustering structure on every view, which is the point we aim to convey.

\begin{figure*}[!htbp]
    \centering
    \subfloat[]{
        \includegraphics[width=0.25\textwidth]{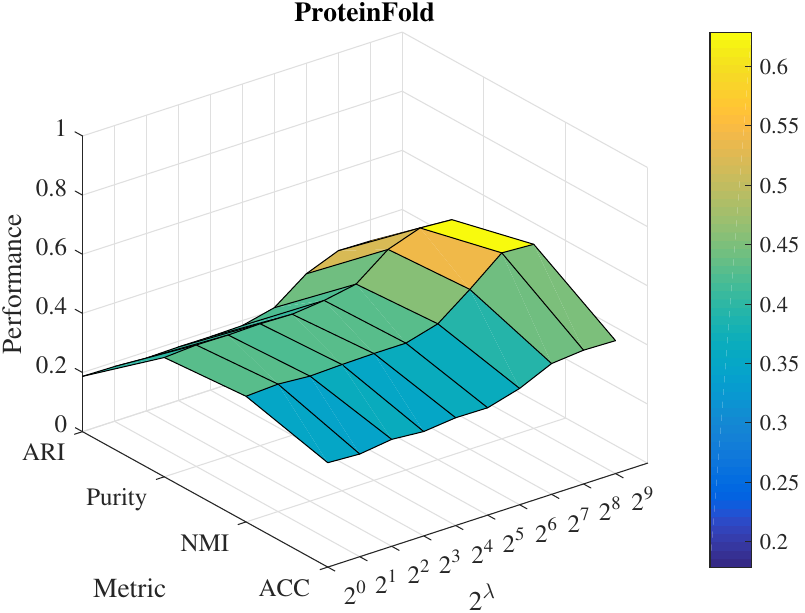} 
        \label{fig7a}
    }
    \hfil
    \subfloat[]{
        \includegraphics[width=0.25\textwidth]{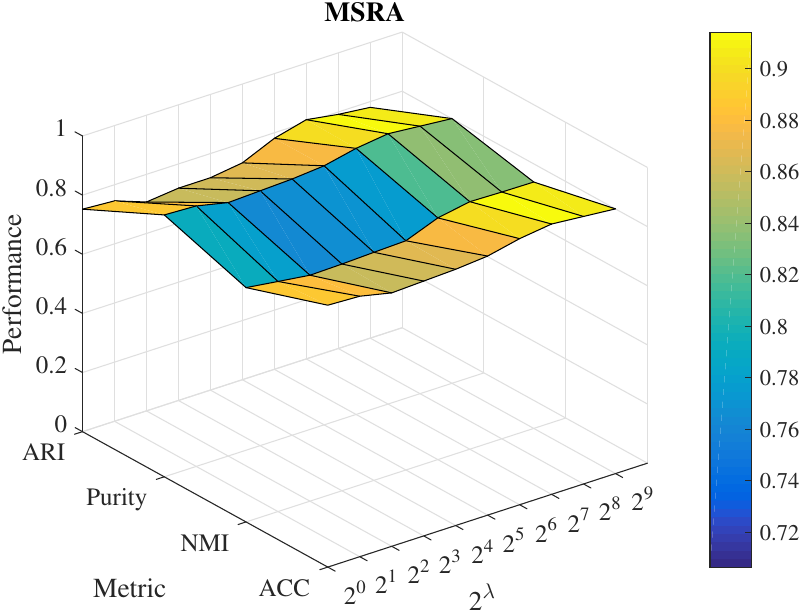} 
        \label{fig7b}
    }
    \hfil
    \subfloat[]{
        \includegraphics[width=0.25\textwidth]{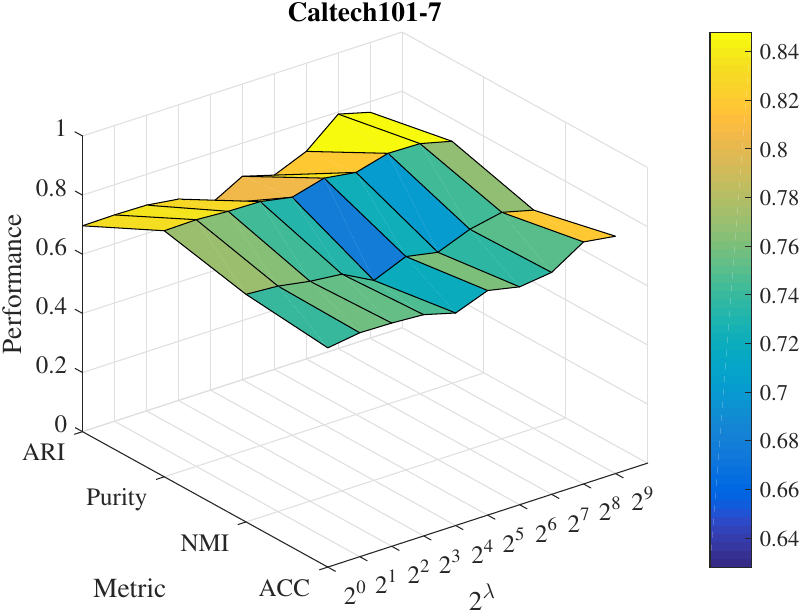} 
        \label{fig7c}
    }
    
    \caption{The Sensitivity of $alpha$ on 3 Datasets. (a) Performance Variation with $\alpha$ on the ProteinFold Datase. (b)Performance Variation with $\alpha$ on the MSRA Dataset. (c) Performance Variation with $\alpha$ on the Caltech101-7 Dataset.}
    \label{fig7}
\end{figure*}

\subsection{Parameter Sensitivity}
In our method, there is only one parameter, $\alpha$ and its range is $\left [ 2^0,2^1,..,2^{9} \right ]$. In Figure \ref{fig7}, we selected three datasets to observe the variation of the ACC and NMI metrics under different values of $\alpha$. The overall change in the surface plot is relatively smooth. On the Cornell dataset, as $\alpha$ increases, all four metrics show an upward trend, and on three datasets, alpha reaches its peak at $2^7$ and $2^8$. This suggests, to some extent, that $\mathbf{H}$ and $\mathbf{G}_v$ need to have a high degree of similarity, but maintaining complete consistency, as in MKKM, would diminish their effectiveness. It is necessary to appropriately allow each view space for learning.

\subsection{Ablation Studies}
As mentioned in the optimization section, we define the objective function as given in Eq. (\ref{eq.14}). Since clarity in clustering results requires $\mathbf{G}_v$ to be sparse, we will present the results of optimizing Eq. (\ref{eq.13}) in this section and compare the different results caused by these two equations. The optimization steps for Eq. (\ref{eq.13}) are similar to those for Eq. Eq. (\ref{eq.14}). And the clustering results based on these two equations are presented in Table \ref{table4}.

Table \ref{table4} presents the clustering effects of using Eq. (\ref{eq.13}) and Eq. (\ref{eq.14}) as objective functions on 9 datasets across 4 metrics. "our-sp" represents Eq. (\ref{eq.14}), while "our-nonsp" represents Eq. (\ref{eq.13}). On the BBC dataset, "our-nonsp" performs better, which may be related to certain characteristics of the dataset. Apart from that, "our-sp" shows superior results on most datasets.

\begin{table}[H]
   		\centering
   		\caption{Results of Ablation Studies\\The Best Results are Highlighted in Red }
   		\label{table4}
   		\resizebox{\linewidth}{!}{
   		\begin{tabular}{cccccc}
   			\toprule
   			Datasets&Methods&ACC&NMI&Purity&ARI\\
   			\midrule
             \multirow{2}{*}{texas}&our-sp&0.5722&\textcolor{red}{0.3667}&0.7380&0.2690\\
   			&our-nonsp&\textcolor{red}{0.5775}&0.3328&\textcolor{red}{0.7487}&\textcolor{red}{0.2908}\\
   			\midrule
            \multirow{2}{*}{cornell}&our-sp&\textcolor{red}{0.6308}&0.4410&0.6974&\textcolor{red}{0.3994}\\
   			&our-nonsp&0.6103&\textcolor{red}{0.4550}&\textcolor{red}{0.7179}&0.3776\\
                \midrule
            \multirow{2}{*}{MSRA}&our-sp&\textcolor{red}{0.9048}&\textcolor{red}{0.8459}&\textcolor{red}{0.9048}&\textcolor{red}{0.7951}\\
   			&our-nonsp&0.8952&0.8038&0.8952&0.7763\\
   			\midrule
            \multirow{2}{*}{Caltech101-7}&our-sp&\textcolor{red}{0.8095}&\textcolor{red}{0.7749}&\textcolor{red}{0.8639}&\textcolor{red}{0.7882}\\
   			&our-nonsp&0.7619&0.7698&0.8322&0.6971\\
   			\midrule
            \multirow{2}{*}{BBCsports}&our-sp&\textcolor{red}{0.9044}&0.7783&\textcolor{red}{0.9044}&0.8027\\
   			&our-nonsp&0.9026&\textcolor{red}{0.7908}&0.9026&\textcolor{red}{0.8099}\\
   			\midrule
   			\multirow{2}{*}{BBC}&our-sp&0.8248&0.7086&0.8248&0.6692\\
   			&our-nonsp&\textcolor{red}{0.8803}&\textcolor{red}{0.7490}&\textcolor{red}{0.8803}&\textcolor{red}{0.7412}\\
   			\midrule
   		\multirow{2}{*}{proteinfold}&our-sp&\textcolor{red}{0.4784}&0.6155&0.5663&\textcolor{red}{0.3289}\\
   			&our-nonsp&0.4755&\textcolor{red}{0.6261}&\textcolor{red}{0.5677}&0.3426\\
   			\midrule
   			\multirow{2}{*}{Caltech101-20}&our-sp&\textcolor{red}{0.4992}&\textcolor{red}{0.6307}&\textcolor{red}{0.8286}&0.3543\\
   			&our-nonsp&0.4908&0.6047&0.8160&\textcolor{red}{0.3657}\\
   			\midrule
   			\multirow{2}{*}{CCV}&our-sp&\textcolor{red}{0.3185}&\textcolor{red}{0.4276}&\textcolor{red}{0.3775}&\textcolor{red}{0.1937}\\
   			&our-nonsp&0.2317&0.1871&0.2553&0.0779\\
   			\bottomrule
   		\end{tabular}
         }   
   	\end{table}

\section{Conclusion} \label{sec6}
In this article, we propose a unified multi-kernel learning approach through matrix factorization for multi-view clustering. Our derivation starts from matrix factorization, and the final objective function is unified into a multi-kernel form. In the derivation process of the objective function, we first remove the non-negativity constraint to expand the learning scope. Additionally, we believe that the clustering structure in a single view is not clear, so we abandon the orthogonality constraint on individual views. Instead, we impose the orthogonality constraint directly on the consensus embedding matrix, allowing room for learning within individual views and obtaining a final accurate structure. Moreover, although the form of the objective function is similar to multi-kernel learning, the proposed method differs from the traditional multi-kernel learning approach. We bypass the direct learning of the optimal kernel to obtain a consensus embedding, which also improves the computational complexity. Finally, comparative experiments on 10 datasets with 9 algorithms have proven the effectiveness of the proposed method.

\section*{Acknowledgment}

The authors would like to thank the editors and anonymous reviewers for their constructive comments. This work is supported by NSFC (No. 12471431, 12231007), Hunan Provincial Natural Science Foundation of China (No. 2023JJ30113) and Guangdong Basic and Applied Basic Research Foundation (No. 2023A1515012342).

\ifCLASSOPTIONcaptionsoff
  \newpage
\fi

\bibliographystyle{IEEEtran}

\bibliography{./bibtex/bib/multiview.bib,./bibtex/bib/matrix_factorization.bib,./bibtex/bib/subspace.bib,./bibtex/bib/MKC.bib}

\end{document}